\RequirePackage{fix-cm} %
\documentclass[10pt,twocolumn,letterpaper]{article}

\PassOptionsToPackage{table}{xcolor}

\usepackage{paper}

\makeatletter
\renewcommand\paragraph{\@startsection{paragraph}{4}{\z@}{6pt \@plus 2pt \@minus 2pt}{-1em}{\normalfont\normalsize\bfseries}}
\makeatother
\definecolor{linkblue}{rgb}{0.21,0.49,0.74}
\usepackage[pagebackref,breaklinks,colorlinks,allcolors=linkblue]{hyperref}

\usepackage{mathtools}
\usepackage{bm}
\usepackage{array}
\usepackage{tabularx}
\usepackage{multirow}
\usepackage{makecell}
\usepackage{nicematrix}
\usepackage{tikz}
\usetikzlibrary{arrows.meta,positioning,calc,patterns}
\usepackage{pgfplots}
\pgfplotsset{compat=1.18}
\usepackage{listings}
\usepackage[most]{tcolorbox}

\definecolor{slBlue}{HTML}{0071BC}
\definecolor{slInk}{HTML}{16191D}
\definecolor{slGray}{HTML}{6E7781}
\definecolor{slRule}{HTML}{D9E1EA}
\definecolor{slTint}{HTML}{F4F8FC}
\definecolor{slLogoBlue}{HTML}{669ED3}
\definecolor{slGreen}{HTML}{67A651}
\definecolor{slRed}{HTML}{D85E5D}
\definecolor{slOrange}{HTML}{F3AA68}
\definecolor{slGain}{HTML}{3F8F3A}

\newcommand{\Bench}{MRBench\xspace} %
\newcommand{\Scissors}{MRScissors\xspace} %
\usepackage{pifont}
\newcommand{\yes}{\textcolor{slGreen}{\ding{51}}}
\newcommand{\no}{\textcolor{slGray}{$\times$}}
\newcommand{\some}{\textcolor{slOrange}{$\triangle$}}

\newcommand{\W}{\textsf{W}\xspace}
\newcommand{\Ss}{\textsf{S}\xspace}
\newcommand{\E}{\textsf{E}\xspace}
\newcommand{\N}{\textsf{N}\xspace}
\newcommand{\nope}{$\varnothing$} %
\newcommand{\ver}[1]{\textit{#1}} %
\newcommand{\suite}[1]{\textbf{\textit{#1}}} %
\newcommand{\Con}{\suite{Concurrency}\xspace}
\newcommand{\Rea}{\suite{Reach}\xspace}

\definecolor{famCon}{HTML}{DE8F8F}\definecolor{famConMid}{HTML}{B86464}\definecolor{famConInk}{HTML}{9B4F4F}
\definecolor{famConLight}{HTML}{EDB3B3}\definecolor{famConTint}{HTML}{FAECEC}
\definecolor{famRea}{HTML}{86A7E3}\definecolor{famReaMid}{HTML}{5D7EC0}\definecolor{famReaInk}{HTML}{46639A}
\definecolor{famReaLight}{HTML}{A9C1EE}\definecolor{famReaTint}{HTML}{EBF1FA}
\colorlet{tabGroup}{famReaTint}
\definecolor{armGlyph}{HTML}{80858C}\definecolor{armGlyphDark}{HTML}{5F646B}
\definecolor{figInk}{HTML}{3A3A3A}\definecolor{figText}{HTML}{555555}\definecolor{figMute}{HTML}{8A8A8A}\definecolor{figAxis}{HTML}{9A9A9A}
\definecolor{figPanel}{HTML}{EDEDED}
\definecolor{verExact}{HTML}{6E7781}
\definecolor{verSurplus}{HTML}{4C8ED1}
\definecolor{verShortfall}{HTML}{D85E5D}
\definecolor{verSwapped}{HTML}{8E6BBE}
\definecolor{verBorrow}{HTML}{4F9A3F}
\definecolor{blkTool}{HTML}{E3EEF9}
\definecolor{blkSim}{HTML}{E6F2E0}

\newcommand{\na}{\textcolor{slGray}{--}} %

\newcommand{\figfont}{\sffamily\fontsize{7}{8.4}\selectfont}
\newcommand{\figsmall}{\sffamily\fontsize{6}{7.2}\selectfont}
\newcommand{\figtiny}{\sffamily\fontsize{5.2}{6.2}\selectfont}

\usetikzlibrary{shapes.callouts,shapes.misc,fit,backgrounds,decorations.pathreplacing}
\tikzset{
  panel/.style={draw=#1, line width=0.9pt, rounded corners=5pt, fill=white},
  card/.style={draw=#1, line width=0.6pt, rounded corners=3pt, fill=white},
  cardhead/.style={fill=#1, text=white, font=\figfont\bfseries, inner sep=2pt, rounded corners=2pt},
  famtag/.style={fill=#1, text=white, font=\figfont\bfseries, inner ysep=2.2pt, inner xsep=4pt, rounded corners=2pt},
  bubble/.style={rectangle callout, callout relative pointer={(-0.22,-0.32)}, draw=slGray, fill=white,
                 rounded corners=2.5pt, font=\figsmall\itshape, inner sep=2.6pt, align=left},
  frame/.style={inner sep=0pt, draw=slRule, line width=0.3pt},
  tbadge/.style={fill=slInk, fill opacity=0.75, text opacity=1, text=white, font=\figtiny, inner sep=1.4pt, rounded corners=1pt, anchor=north west},
  tbl/.style={draw=slGray, fill=slTint, rounded corners=1pt, line width=0.4pt},
  used/.style={circle, draw=slBlue, fill=slBlue, text=white, inner sep=0pt, minimum size=3.6mm, font=\figtiny\bfseries},
  idle/.style={circle, draw=slGray, fill=white, text=slGray, inner sep=0pt, minimum size=3.6mm, font=\figtiny\bfseries, line width=0.6pt},
  gone/.style={circle, draw=slRed, dashed, fill=white, text=slRed, inner sep=0pt, minimum size=3.6mm, font=\figtiny\bfseries, line width=0.6pt},
  swapin/.style={circle, draw=verSwapped, fill=verSwapped, text=white, inner sep=0pt, minimum size=3.6mm, font=\figtiny\bfseries},
  chan/.style={line width=1.5pt, draw=slOrange!85!black, line cap=round},
  chanmiss/.style={line width=1.5pt, draw=slRed, dash pattern=on 1.4pt off 1.2pt, line cap=round},
  prop/.style={line width=1.5pt, draw=verBorrow, line cap=round},
  decoy/.style={line width=1.5pt, draw=slRed!75, line cap=round},
  blk/.style={draw=#1!60!black, fill=#1, rounded corners=3pt, font=\figfont, align=center, inner sep=3pt, line width=0.5pt},
  blktitle/.style={font=\figfont\bfseries, text=slInk},
  arrow/.style={-{Stealth[length=4pt,width=3.5pt]}, line width=0.6pt, draw=slInk!80},
  note/.style={font=\figtiny, text=slGray, align=center},
}

\title{MRBench: Benchmarking Physical Necessity Reasoning in Heterogeneous Multi-Robot Manipulation}

\author{%
Yanzhe Chen$^{*}$\quad Zhijun Cao$^{*}$\quad Ruihe An$^{*}$\quad Chang Gong$^{*}$\quad Mike Zheng Shou$^{\dagger}$\\
Show Lab, National University of Singapore\\
{\small $^{*}$Equal contribution\qquad $^{\dagger}$Corresponding author}%
}

\begin{document}
\twocolumn[{%
\renewcommand\twocolumn[1][]{#1}%
\maketitle
\begin{center}
  \centering
  \captionsetup{type=figure}
  \includegraphics[width=\textwidth]{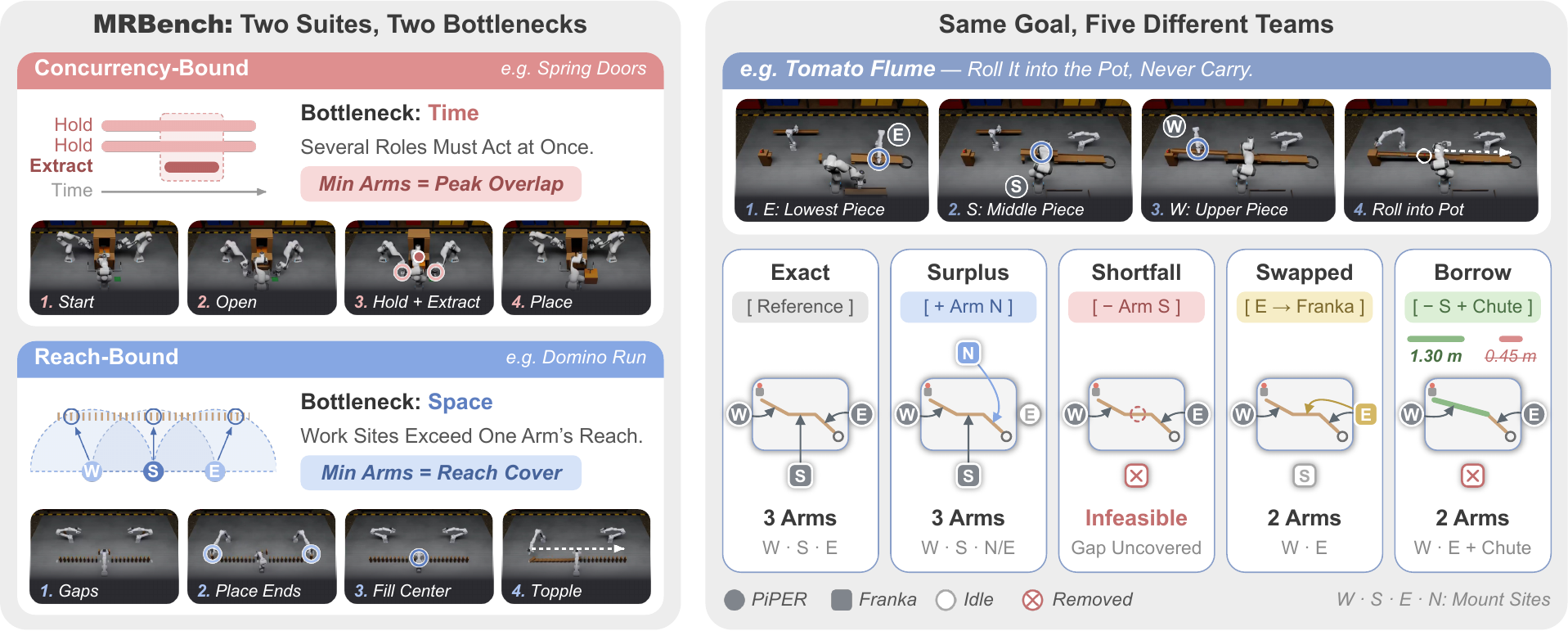}
  \captionof{figure}{
\textbf{Physical necessity depends on task demands and available resources.}
Left: \textbf{Two physical bottlenecks}: \Con tasks require simultaneous actions; \Rea tasks require spatial coverage.
Right: \textbf{Same goal, different teams}: three arms in \ver{exact}; a redundant arm in \ver{surplus}; infeasible in \ver{shortfall}; two arms with a different robot in \ver{swapped}, or a passive substitute in \ver{borrow}.
  }
  \label{fig:teaser}
\end{center}
\vspace{0.6em}%
}]

\begin{abstract}
Recent advances in multi-robot manipulation have shown how robots can collaborate to accomplish complex tasks.
However, successful coordination does not reveal which robots are actually needed.
The same task may require different teams as robot capabilities change, resources become unavailable, or objects replace robotic support.
We argue that reliable multi-robot manipulation requires \textbf{physical necessity reasoning}: understanding what a task demands, which resources can fulfill those demands, and which are truly necessary.
We introduce \textbf{\Bench}, a benchmark for physical necessity reasoning in \textbf{heterogeneous multi-robot manipulation}. It comprises eight tasks with distinct concurrency and reach constraints, each evaluated under five configurations that change the required resources while preserving the task goal.
Our evaluation reveals that existing coordination agents perform well with reference teams but struggle when robot embodiments change or passive objects can replace robotic support.
To address this gap, we propose \textbf{\Scissors}, which separates task requirements from measured physical capabilities and identifies minimal feasible teams through constrained search.
Across two model backbones, \Scissors achieves 85.4\% and 91.2\% team-selection accuracy, outperforming the strongest baselines by 10.4 and 9.5 points, respectively.
Together, these findings expose a critical gap between \textbf{coordinating robots and understanding their physical necessity}.

\vspace{0.2em}
{\upshape\noindent
\textbf{Website:}~\href{https://showlab.github.io/MRBench/}{https://showlab.github.io/MRBench}
\par}
\end{abstract}

\section{Introduction}
\label{sec:intro}

\begin{quote}
\vspace{-0.5em}
\textit{``Plurality is never to be posited without necessity."} --- William of Ockham
\end{quote}
Recent advances in foundation-model agents have enabled robots to coordinate manipulation tasks through task decomposition, collaborative planning, and shared execution~\cite{mandi2024roco,kannan2024smartllm,chen2024hmas,chen2025emos,kang2026coenv}.
Yet \textbf{successful coordination does not reveal which robots are actually needed}.
In practical workcells, robots may become unavailable, be replaced by different models, or have their roles taken over by simple fixtures.
This raises a question beyond coordination:
\emph{Which physical capabilities does a task truly require, and which robots or objects are necessary to provide them?}

The answer depends on the task's physical constraints rather than robot count alone.
Some tasks require several independent actions at once, while others span locations beyond a single arm's reach (\cref{fig:teaser}A).
Crucially, \textbf{required physical roles do not always require additional robots}.
As shown in \cref{fig:teaser}B, the same tomato-flume task needs three arms under its reference configuration, becomes infeasible when a key arm is removed, but can be completed with two when a different arm reaches farther or a suitable chute replaces the missing support.
The task goal remains unchanged, yet its minimal feasible team changes.
This motivates \textbf{physical necessity reasoning}: determining what a task demands, what the available resources can provide, and which combination is minimally sufficient.

Classical multi-robot task allocation studies capability-based team formation~\cite{gerkey2004taxonomy,korsah2013taxonomy}, while recent embodied agents extend robot selection and coordination to heterogeneous teams~\cite{kannan2024smartllm,chen2025emos,kang2025vikir,zhang2025toolroco}.
However, existing benchmarks largely evaluate coordination ability with fixed teams or selection from available robots~\cite{mandi2024roco,qin2025robofactory}.
They do not systematically test whether agents can infer task requirements, assess physical capabilities, and recognize when passive objects can replace robotic roles.
\textbf{A successful execution proves that a team is sufficient, not that it is necessary}.

To investigate this gap, we introduce \textbf{\Bench}, a benchmark for physical necessity reasoning in heterogeneous multi-robot manipulation.
\Bench comprises eight tasks spanning two sources of multi-arm necessity: \Con and \Rea constraints (\cref{fig:tasks}).
Each task is evaluated under five configurations---\ver{exact}, \ver{surplus}, \ver{shortfall}, \ver{swapped}, and \ver{borrow}---that preserve the goal while changing the available physical resources.
Ground-truth minimal teams are computed from geometric constraints and exact search, with feasible solutions validated in Isaac Sim~\cite{mittal2025isaaclab}.
This design tests whether agents can distinguish necessary, redundant, missing, and replaceable robotic support.

Our evaluation reveals a striking gap: \textbf{agents that coordinate reference teams successfully often struggle when physical resources change}.
Existing coordination methods achieve high accuracy under reference configurations, yet degrade substantially when robot embodiments are swapped or passive objects can replace robotic roles.
To address this gap, we propose \textbf{\Scissors}, which separates \emph{task requirements} from \emph{measured physical capabilities}.
A team-independent task representation identifies the required jobs, physical tools determine which robots or objects can perform them, and constrained search produces a minimal feasible team with a verifiable assignment.

Across two model backbones, \Scissors achieves 85.4\% and 91.2\% team-selection accuracy, outperforming the strongest baselines by 10.4 and 9.5 points.
The same solver operating on model-generated capabilities trails by 19.2 and 13.1 points, highlighting the importance of physical measurement.
Moreover, \Scissors' plans achieve over twice the execution success rate of competing methods when the selected team is correct.
Nevertheless, passive substitution remains challenging, revealing a persistent gap between \textbf{coordinating robots and understanding their physical necessity}.
Our contributions are threefold:
\begin{itemize}
\item \textbf{Empirical Insight.} We reveal a gap between multi-robot coordination and physical necessity reasoning as robot capabilities and environmental support change.
\item \textbf{\Bench.} We introduce a benchmark evaluating feasibility, minimality, and functional substitution across heterogeneous configurations under unchanged task goals.
\item \textbf{\Scissors.} We propose a framework that separates task requirements from measured physical capabilities, improving team selection and downstream execution.
\end{itemize}

\begin{table}[t]
  \caption{\textbf{Closest work.} Compared by what is decided on the \emph{team}, what an \emph{answer} says and how it is \emph{evaluated} (\emph{Hetero.}: heterogeneous). \yes: reported; \some: narrower evidence; \no: absent.}
  \label{tab:related}
  \centering
  \footnotesize
  \setlength{\tabcolsep}{0.3pt}
  \newcommand{\hd}[1]{{\scriptsize\makecell{#1}}}
  \begin{NiceTabular*}{\linewidth}{@{\extracolsep{\fill}}l*{6}{w{c}{28pt}}@{}}[cell-space-limits=1.2pt]
    \CodeBefore
      \rowcolor{tabGroup}{3,10}
    \Body
    \toprule
    \Block[l]{2-1}{Work} & \Block{1-2}{Team} & & \Block{1-2}{Answer} & & \Block{1-2}{Evaluation} & \\
    \cmidrule(lr){2-3} \cmidrule(lr){4-5} \cmidrule(lr){6-7}
    & \hd{Hetero.\\arms} & \hd{Selects\\team} & \hd{Infeasible\\answer} & \hd{Passive\\substitute} & \hd{Matched\\changes} & \hd{Physics\\execution} \\
    \midrule
    \Block[l]{1-7}{\textbf{Foundation-model agents for robot teams}} \\
    RoCo~\cite{mandi2024roco} & \yes  & \no   & \no   & \no   & \no   & \yes \\
    SMART-LLM~\cite{kannan2024smartllm} & \no   & \yes  & \some & \no   & \no   & \yes \\
    HMAS-2~\cite{chen2024hmas} & \no   & \no   & \no   & \no   & \no   & \yes \\
    EMOS~\cite{chen2025emos} & \yes  & \some & \no   & \no   & \no   & \no \\
    CoEnv~\cite{kang2026coenv} & \yes  & \no   & \no   & \no   & \no   & \yes \\
    Tool-RoCo~\cite{zhang2025toolroco} & \yes  & \yes  & \no   & \no   & \no   & \yes \\
    \Block[l]{1-7}{\textbf{Multi-robot benchmarks}} \\
    RoboFactory~\cite{qin2025robofactory} & \no   & \no   & \no   & \no   & \no   & \yes \\
    VIKI-Bench~\cite{kang2025vikir} & \yes  & \yes  & \no   & \some & \no   & \no \\
    \midrule
    \textbf{\Bench (ours)} & \yes  & \yes  & \yes  & \yes  & \yes  & \yes \\
    \bottomrule
  \end{NiceTabular*}
\end{table}

\section{Related Work}
\label{sec:related}

\paragraph{Benchmarks for multi-robot tasks.}
Benchmarks have grown from one arm~\cite{liu2023libero} to two~\cite{zhao2023aloha,grotz2024peract2,mu2025robotwin,chen2025robotwin2,julg2026duobench} and several, with tasks built around coordination: arms that share a table (RoCoBench~\cite{mandi2024roco}), compositional constraints (RoboFactory~\cite{qin2025robofactory}), humanoid teams (CoHuB~\cite{park2026cohub}) and a human partner (PARTNR~\cite{chang2025partnr}); other suites vary scenes, instructions, models or harnesses on a fixed task~\cite{zhou2025liberopro,fei2026liberoplus,fu2026capx,chen2026showharness,yang2026robotworld,espinola2026libero10}.
Their \emph{tightly coupled} tasks, which no single robot can finish alone for lack of hands or of reach, come with the team, so a successful run shows that the team suffices, not that each arm is needed.
VIKI-Bench~\cite{kang2025vikir} and GroundAct~\cite{wang2025groundact} do ask which robots a task needs, but have no infeasible task; to our knowledge, no benchmark asks for the fewest arms of a tightly coupled task and accepts that no team exists.
\Bench does both (\cref{tab:related}): five configurations per task change the answer under one instruction, as in contrast sets~\cite{gardner2020contrast}, and 14 of its 40 versions have no feasible team, like the unanswerable questions of SQuAD~2.0~\cite{rajpurkar2018squad2}.

\paragraph{Planners and agents for multi-robot tasks.}
Most multi-robot manipulation work plans for a team that the task supplies~\cite{smith2012dualarm,sugar2002mobile,farivarnejad2022transport,tung2021mart,ha2020multiarm,lai2025roboballet,wang2025gaudp,he2025latenttom,dong2026mimicd,zhang2026mavla,doshi2026chorus,gao2024coohoi,lionar2026teamhoi}, often on tightly coupled tasks, and work that does choose arms starts from full models or stated requirements~\cite{chen2020soma,hartmann2026wrap,holladay2021forceful,xu2023robotool,marvel2018multirobot,gerkey2004taxonomy,korsah2013taxonomy,rudolph2021riskadaptive,martin2026clustering}.
Foundation-model agents instead decide at run time what each robot does: RoCo~\cite{mandi2024roco} lets arm agents talk until they agree on subtasks and waypoints, HMAS-2~\cite{chen2024hmas} has a central model propose actions that each robot checks, and SMART-LLM~\cite{kannan2024smartllm} forms coalitions from skills and may leave robots unassigned or decline to plan; others coordinate by messages, shared programs, or planners and verifiers~\cite{zhang2024coela,liu2025coherent,chen2025emos,kang2026coenv,zhang2025toolroco,liu2023llmp,zhang2025lammap,rivera2025fleet,curtis2024proc3s,wang2024llm3,kambhampati2024llmmodulo}, and single agents learn to ask for help or refuse~\cite{ahn2022saycan,ren2023knowno,hsieh2025dowhat,cheng2026feasibility}.
These works score success, steps, tokens or utilization, but not how many arms a task needs or how many model sessions deciding takes.
\Scissors addresses both by splitting the decision: a model blind to the arms writes the jobs the task needs, code measures what every arm can do, and an exact search returns the fewest arms with a certificate of why each is needed, all in about four sessions.

\section{\texorpdfstring{\Bench}{MRBench}}
\label{sec:bench}

\begin{figure*}[t]
  \centering
  \includegraphics[width=\textwidth]{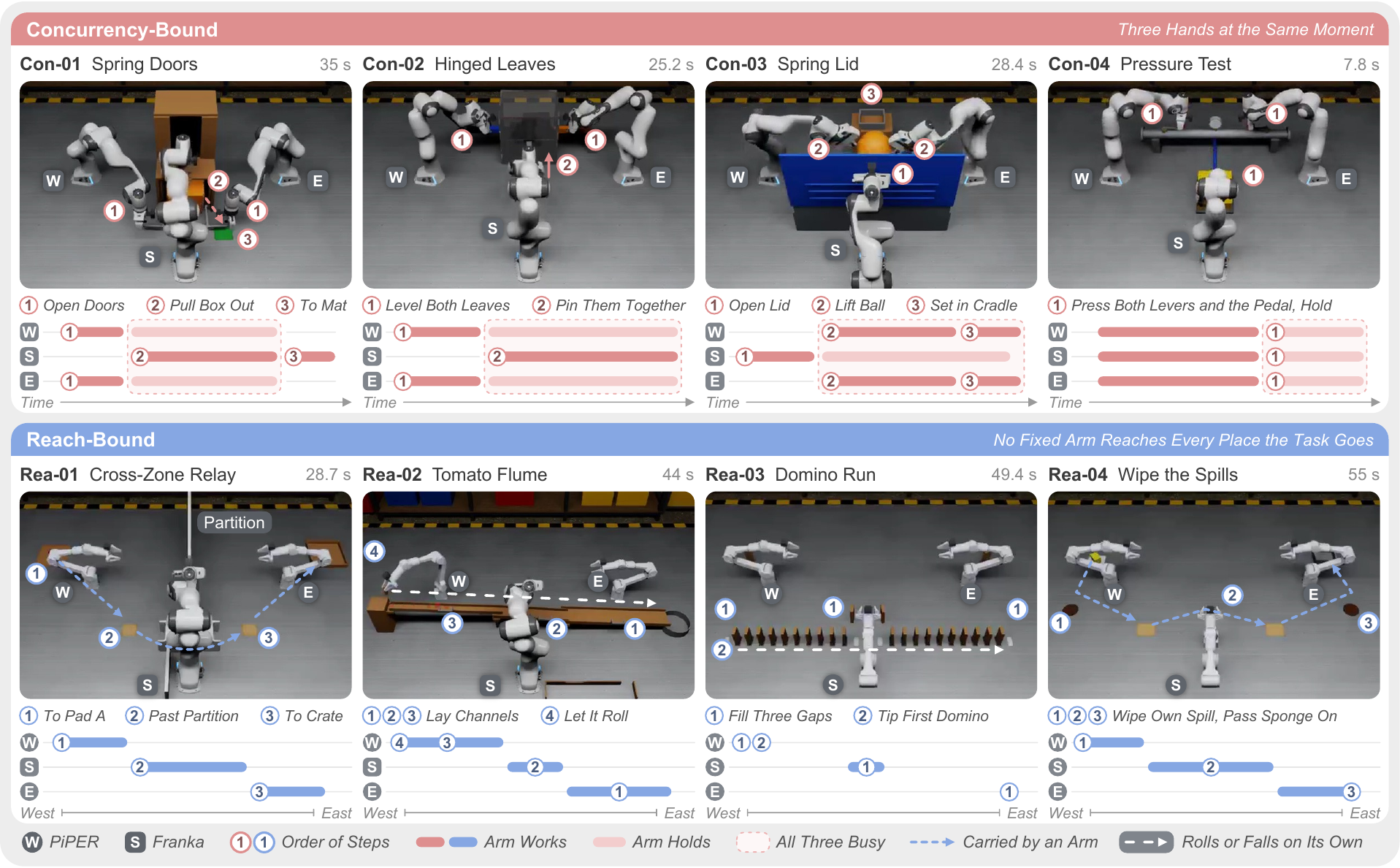}
  \caption{\textbf{Overview of the eight tasks in \Bench.} Each is shown at a key moment of its scripted reference (\ver{exact}, easy level), with its steps in order and when (\Con) or where (\Rea) each arm works. The instruction names no arm.}
  \label{fig:tasks}
\end{figure*}

\Bench asks which arms, or which passive objects in their place, a task truly needs, when the goal stays fixed and the minimal feasible team changes.
Eight tasks (\cref{fig:tasks}) in five configurations give 40 versions and 480 scenes.

\subsection{Physical necessity as a minimal support set}
\label{sec:problem}

A task is a set of \emph{jobs} done by \emph{doers}; it needs several arms when jobs at different places must happen at once (\Con) or lie beyond one arm's reach (\Rea).

\paragraph{Jobs.}
Each job $j\in J$ has \emph{sites} $S_j$, the poses where a hand must act, and a \emph{failure signal} that the simulator reports when it is not done.
An \emph{act} job needs one doer that reaches every site; a \emph{carry} job moves a body from $s^0_j$ to $s^1_j$, by one doer or a chain of doers that meet at hand-over points.
The jobs of a group $g\in\mathcal{G}\subseteq 2^J$ must happen at once, such as a door held open while a box is drawn out.

\paragraph{Doers.}
The doers are the arms $A$, each a robot model at a fixed place, and the \emph{passive objects} $P$ on the table, which can take over a job from an arm.
An arm $a$ can do an act job $j$, written $c(a,j)=1$, when it reaches every site with margin and its fingers open wide enough:
\begin{equation}
  \rho_a(s)\ge\delta_j \quad \forall s\in S_j
  \qquad\text{and}\qquad
  w_j\le o_a ,
  \label{eq:cap}
\end{equation}
where $\rho_a(s)$ is the inverse-kinematics (IK) margin at $s$, $\delta_j$ the margin required (2--8.5\,cm), $w_j$ the grasp width, $o_a$ the finger opening.
A chain of arms can do a carry job when its ends reach $s^0_j$ and $s^1_j$ and consecutive arms share a hand-over point, all with margin.
An object $p$ can do $j$, $c(p,j)=1$, when a measured property $\phi_j(p)$ lies in $\Theta_j$ (a weight taller than a lever's bar); an arm must place it (job $j_p$).

\paragraph{Minimal support sets.}
A team $T\subseteq A$ is \emph{feasible} if, for some $U\subseteq P$, an assignment $\sigma$ gives every job $j\in J_U = J\cup\{j_p: p\in U\}$ doers $\sigma(j)\subseteq U\cup T$ such that
\begin{equation}
  \begin{aligned}
    &c(\sigma(j),j)=1 && \forall j\in J_U,\\
    &\sigma(j)\cap\sigma(j')=\varnothing && \forall g\in\mathcal{G},\ j\neq j'\in g .
  \end{aligned}
  \label{eq:feasible}
\end{equation}
\emph{Minimal support sets} (MSS) are the smallest feasible teams,
\begin{equation}
  \mathrm{MSS}(A,P)=\operatorname*{arg\,min}_{T\subseteq A\ \text{feasible}} |T| ,
  \label{eq:mss}
\end{equation}
and any one of them is a right answer; the task is \emph{infeasible} when there is none, that is, when $A$ itself is not feasible.
An answer thus settles \emph{minimality} (who is needed), \emph{feasibility} (whether the task can be done) and \emph{functional substitution} (what can stand in for an arm).

\subsection{Task design}
\label{sec:tasks}
\label{sec:versions}

\paragraph{Concurrency and Reach.}
The two conditions of \cref{eq:feasible} give two suites of four tasks.
The \Con tasks need three hands at once, one more than a two-handed robot has: two bodies held while a third hand works (Con-01, Con-02), one held while two work (Con-03), or three held together (Con-04).
In the \Rea tasks no two arms act at once, but no fixed arm reaches every place: bodies pass between arms by being set down, or each arm works where only it reaches (\cref{fig:tasks}).
We leave out loads too heavy for one arm: privileged grasps do not model grip or payload.

\paragraph{Five configurations.}
They change only a task's doers; each but the reference defeats a shortcut that skips measuring.
(i)~The reference \ver{exact} needs all three arms.
(ii)~\ver{surplus} adds a redundant Franka at \N; it defeats \emph{staffing by presence}.
(iii)~\ver{shortfall} removes a needed arm, so no team is feasible; the answer must name the job without a doer; it defeats \emph{presumed feasibility}.
(iv)~\ver{swapped} trades one arm for another embodiment, which can leave every place staffed yet the task infeasible, or let two arms do it; it defeats \emph{roles by place}.
(v)~\ver{borrow} swaps a needed arm for a prop that can do its job and a distractor that cannot, alike but for one measured property; it defeats \emph{substitution by appearance}.

\paragraph{Design rules.}
The instruction gives only the goal: no arm, role, step or tool.
Easy and hard levels differ in one parameter; success is judged by state predicates, not by a method's claims; physics is tuned on ten other tasks and a prototype.

\subsection{Benchmark construction}
\label{sec:gold}

\paragraph{Workcell and scenes.}
One table has three fixed places, west (\W) and east (\E) 1.4\,m apart with south (\Ss) between, and a fourth, north (\N), for the \ver{surplus} arm; each takes a Franka, a PiPER or an SO-101.
We simulate in Isaac Sim~4.5 with Isaac Lab~2.1~\cite{mittal2025isaaclab}; grasps are privileged but refused on a body wider than the open fingers unless a second hand presses its far side.
Each version has 12 scenes (two levels, six seeds): the seed picks the room, recolors the task's objects and adds three everyday items.

\paragraph{Ground truth is computed.}
The ground-truth (gold) answer of a scene comes from IK with margins (\cref{eq:cap}) and an exact search for the smallest feasible teams (\cref{eq:mss}).
In Isaac Sim the scripted references succeed in all 312 feasible scenes and the distractors fail as designed in all 96 \ver{borrow} scenes; the solver proves infeasibility, and every two-arm attempt in Isaac reports the job left undone.

\paragraph{Audits.}
Each claim that an arm is necessary holds under IK with margins or, where IK cannot decide, in Isaac.
Fixed rules, which never measure, do well per version but not per task: the best (count the arms, take the larger object, call any team with an SO-101 infeasible) gets 25 of the 96 task-scenes right, so task-level accuracy is the headline.

\subsection{Protocol and metrics}
\label{sec:protocol}

The protocol removes four conveniences that multi-robot benchmarks offer and real workcells do not.
(i)~No team is given: a method gets the goal, each arm's model, place and orientation, and the objects, but no role, skill list or reach.
(ii)~Feasibility is not promised: 14 of the 40 versions are infeasible, and ``infeasible: these jobs cannot all have a doer'' is then the answer.
(iii)~Names do not decide: a prop and its distractor look alike, and masses are given only within a factor of 0.75--1.3, so fitness must be measured.
(iv)~Failure is not proof: a failed run's \emph{failure report} names the jobs whose signals fired, and only such a signal, not a missed grasp, counts against a smaller team; reading any failure as a need for more arms is a fifth shortcut, \emph{failure as proof}.

\paragraph{Tools and settings.}
Every method has the same senses, measuring tools (\texttt{measure}, \texttt{heft}, \texttt{can\_reach}, \texttt{who\_can}, \texttt{meet}, \texttt{check\_hold}) and \texttt{probe}, moves arms only through one program API (\texttt{check\_program}, \texttt{try\_in\_sim}) and submits one form: feasibility, arms used and idle, objects used and rejected with evidence, steps, and what is missing.
It submits without a run in the \emph{one-shot} setting; under the \emph{equal budget} it has 2 probes and 3 runs per scene and may revise after a failure.

\paragraph{Metrics.}
\label{sec:metrics}
\emph{Team-selection accuracy} scores each scene against the gold answer: right only if feasibility, arms and objects all match, counting every arm that moved, so a method cannot declare two arms and run three.
\emph{Task-level accuracy} is the share of the 96 task-scenes (a task at one level and seed) with all five configurations right.
We also report the success rate (SR) on feasible scenes, \emph{MSS-SR} (right team, successful run), the cost and error types.

\section{Baselines}
\label{sec:baselines}

Prior methods answer neighboring questions: how a given team cooperates~\cite{mandi2024roco,chen2024hmas}, or which robots get which subtasks~\cite{kannan2024smartllm,chen2025emos}.
We adapt each by one rule, \emph{change what a method takes in and gives out, never how it reasons}: all get one output form, passive candidates and the same tools, split by role, and differ in who decides, where they learn what an arm can do, and what they revise after failure.

\paragraph{References and a solver.}
\label{sec:refs}
\emph{Oracle-script} replays the gold team's scripted reference as the ceiling for SR; \emph{Oracle-team} hands the gold team to the executor, so it fails only in execution.
In \emph{LLM + solver}, as in LLM+P~\cite{liu2023llmp} and FLEET~\cite{rivera2025fleet}, the model translates the scene into jobs, groups and doers (\cref{sec:problem}), and exact search returns the fewest arms or ``infeasible'', so its errors are translation errors.

\paragraph{Prior methods, adapted.}
\label{sec:prior}
We adapt one method per way of organizing a team: a fixed pipeline (SMART-LLM), per-arm dialogue (RoCo) and a central proposal with arm vetoes (HMAS-2).
EMOS~\cite{chen2025emos} allocates centrally from robot resumes, Tool-RoCo~\cite{zhang2025toolroco} activates arm agents centrally or lets them join on their own, and CoEnv~\cite{kang2026coenv} coordinates a given team, so each falls under one of these.
\emph{SMART-LLM-style}~\cite{kannan2024smartllm} runs its three stages, decomposition, coalition formation and allocation, as three sessions and keeps its released instruction to use fewest robots; coalitions are formed by querying the arms, and a subtask without one makes the task infeasible.
\emph{RoCo-style}~\cite{mandi2024roco} gives each arm its own model, which sees the fixed cameras and its own wrist camera, queries only its own kinematics and may decline to take part; the arms talk in turn, up to three rounds, until all agree on a plan, which may declare the task infeasible.
\emph{HMAS-2-style}~\cite{chen2024hmas} has a central model, which sees the scene, not the arms' kinematics, propose team, idle arms, objects and feasibility; each arm accepts or vetoes its part with its own \texttt{can\_reach}, for up to three rounds.

\paragraph{Fair comparison.}
\label{sec:executor}
Every method hands its answer to one shared executor, after CaP-X~\cite{fu2026capx}: a fresh session of the same model receives the whole answer (the arms used and idle, the objects, the steps and the constraints the method names), writes a program for the declared arms, checks it with \texttt{check\_program} and runs it with \texttt{try\_in\_sim}.
Oracle-team hands it the gold team and objects without steps, and a feasible scene answered as infeasible counts as failed.
Under the equal budget, every method receives the same failure report after a failed run and may change the team.
On each backbone all methods share information, tools, output form, budget and tuning effort.

\section{\texorpdfstring{\Scissors}{MRScissors}}
\label{sec:method}

\begin{figure}[!ht] %
  \centering
  \includegraphics[width=\columnwidth]{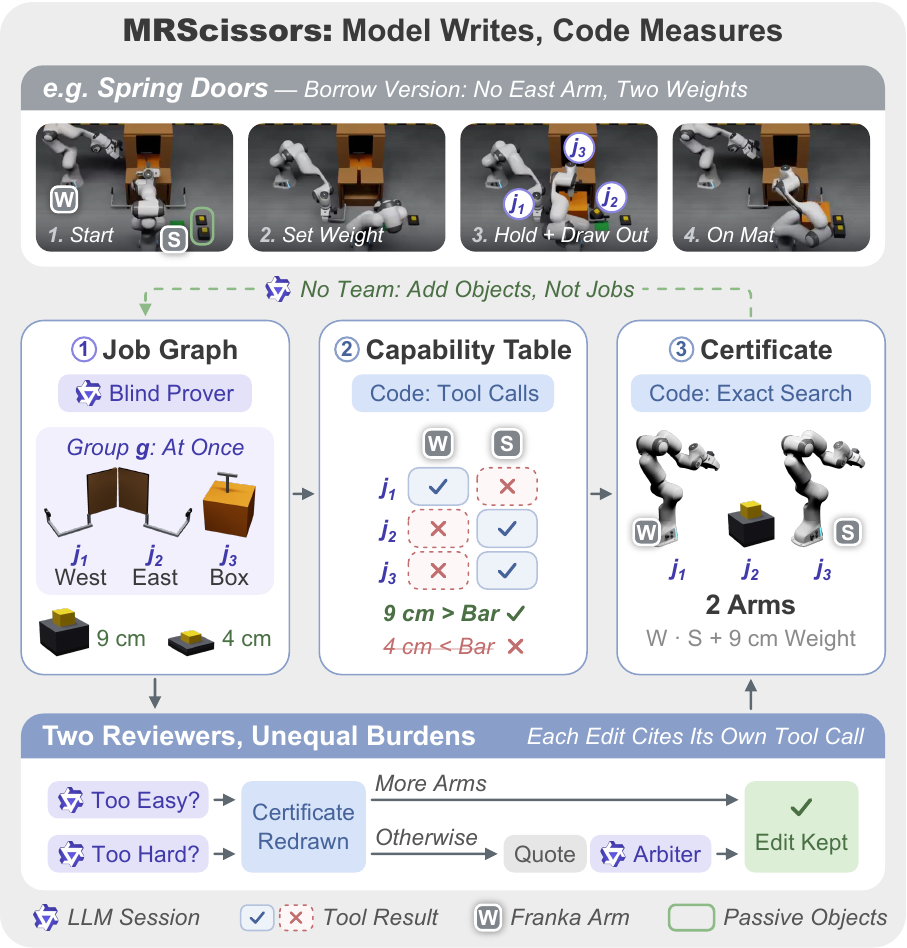}
  \caption{\textbf{Overview of \Scissors.} A blind prover writes the job graph, code measures the capability table, exact search returns the certificate, and two reviewers check the graph.}
  \label{fig:method}
\end{figure}

\begin{table*}[t]
  \caption{\textbf{Team selection and task success on \Bench.} Team-selection accuracy (\%) is reported by configuration, by task suite (\suite{Con.}: \Con; \suite{Rea.}: \Rea), over all 480 episodes and at the task level (Task: 96 task-scenes). SR and MSS-SR (\%) are on the 312 feasible episodes. Cost per episode: arms named, thousands of tokens and model sessions. Bold: best per backbone.}
  \label{tab:main}
  \centering
  \small
  \setlength{\tabcolsep}{1.5pt}
  \begin{NiceTabular*}{\textwidth}{@{\extracolsep{\fill}}l*{5}{w{c}{33pt}}*{2}{w{c}{22pt}}*{3}{w{c}{22pt}}w{c}{34pt}w{c}{22pt}w{c}{28pt}w{c}{20pt}@{}}[cell-space-limits=1.3pt]
    \CodeBefore
      \rowcolor{tabGroup}{3,9}
    \Body
    \toprule
    \Block[l]{2-1}{Method} & \Block{1-5}{By configuration} & & & & & \Block{1-2}{By suite} & & \Block{2-1}{All} & \Block{2-1}{Task} & \Block{1-2}{Task success} & & \Block{1-3}{Cost per episode} & & \\
    \cmidrule(lr){2-6} \cmidrule(lr){7-8} \cmidrule(lr){11-12} \cmidrule(lr){13-15}
    & \ver{exact} & \ver{surplus} & \ver{shortfall} & \ver{swapped} & \ver{borrow} & \suite{Con.} & \suite{Rea.} & & & SR & MSS-SR & Arms & Tokens & Sess. \\
    \midrule
    \Block[l]{1-15}{\textbf{Qwen3.8-Max}} \\
    LLM + solver           & 82.3 & 87.5 & 60.4 & 50.0 & 51.0 & 62.1 & 70.4 & 66.2 & 10.4 & 15.4 & 15.1 & 2.40 & 327.8 & 1.0 \\
    SMART-LLM-style        & 94.8 & 80.2 & 85.4 & 53.1 & 49.0 & 73.8 & 71.2 & 72.5 & 13.5 & 19.6 & 18.3 & 2.52 & 640.6 & 3.0 \\
    RoCo-style             & 85.4 & 85.4 & 52.1 & 46.9 & 35.4 & 57.9 & 64.2 & 61.0 & 2.1 & 13.1 & 12.8 & 2.43 & 1155.5 & 6.1 \\
    HMAS-2-style           & \textbf{95.8} & 81.2 & 86.5 & 64.6 & 46.9 & 75.8 & 74.2 & 75.0 & 17.7 & 18.6 & 17.0 & 2.61 & 764.9 & 7.8 \\
    \midrule
    \textbf{\Scissors (ours)} & 93.8 & \textbf{96.9} & \textbf{93.8} & \textbf{83.3} & \textbf{59.4} & \textbf{87.5} & \textbf{83.3} & \textbf{85.4} & \textbf{43.8} & \textbf{46.8} & \textbf{46.8} & 2.51 & 421.5 & 3.8 \\
    \midrule
    \Block[l]{1-15}{\textbf{Claude Sonnet 5.5}} \\
    LLM + solver           & 94.8 & 79.2 & 92.7 & 68.8 & 55.2 & 77.1 & 79.2 & 78.1 & 18.8 & 19.2 & 18.9 & 2.46 & 282.2 & 1.0 \\
    SMART-LLM-style        & 96.9 & 82.3 & 94.8 & 75.0 & 59.4 & 82.5 & 80.8 & 81.7 & 32.3 & 23.4 & 21.8 & 2.52 & 610.0 & 3.0 \\
    RoCo-style             & 88.5 & 76.0 & 91.7 & 71.9 & 53.1 & 77.9 & 74.6 & 76.2 & 21.9 & 16.7 & 16.7 & 2.46 & 783.8 & 3.6 \\
    HMAS-2-style           & 95.8 & 77.1 & 93.8 & 71.9 & 55.2 & 76.7 & 80.8 & 78.8 & 32.3 & 23.1 & 20.8 & 2.63 & 765.7 & 6.1 \\
    \midrule
    \textbf{\Scissors (ours)} & \textbf{99.0} & \textbf{96.9} & \textbf{97.9} & \textbf{95.8} & \textbf{66.7} & \textbf{91.7} & \textbf{90.8} & \textbf{91.2} & \textbf{59.4} & \textbf{51.6} & \textbf{51.6} & 2.55 & 452.5 & 3.9 \\
    \bottomrule
  \end{NiceTabular*}
\end{table*}

\Scissors separates the two inputs of \cref{eq:feasible}: the model writes what the task needs (the jobs $J$, the groups $\mathcal{G}$ and possible stand-ins) as a \emph{job graph}, code measures who can do it ($c$) into a \emph{capability table}, and an exact search over that table returns the fewest arms (\cref{fig:method}).
It uses only the tools of \cref{sec:protocol} and answers in one pass.

\paragraph{Job graph from a blind prover.}
One model session, the \emph{prover}, writes the instance of \cref{sec:problem} from the goal and the scene: the jobs $\hat{J}$ with their sites, the groups $\hat{\mathcal{G}}$, and each object that could take over a job from an arm, with bounds on its tool readings.
The prover is \emph{blind}: it knows how wide a hand opens, but not which arms are on the table, where they stand or what they reach, and its only tools are its senses, \texttt{measure} and \texttt{heft}.
It thus cannot write the task to suit the arms, and gives look-alikes the same bounds.

\paragraph{Measured capability table.}
Code first makes the graph concrete: a body the prover judges too wide for one hand becomes two acts in one group, and a carry is checked where the gripped part arrives.
It then fills the table $\hat{c}$ of \cref{eq:cap} from tool results alone: \texttt{who\_can} for every arm at every site, \texttt{meet} for every pair of arms on a carry, and \texttt{measure} for every bound of every object.
A reach cell reads \emph{can}, \emph{cannot} or, when the arm falls short by at most 2\,cm (the tool--simulator gap), \emph{near}; an object passes only if all its bounds hold.
Each cell keeps its evidence.

\paragraph{Exact search and certificate.}
On $\hat{J}$, $\hat{\mathcal{G}}$ and $\hat{c}$, code solves \cref{eq:mss} with the search of LLM + solver (\cref{sec:refs}): it tries subsets of arms from one arm up and returns every smallest team in which each job has a doer and the jobs of a group distinct doers, using an object only where hands fall short.
The two methods thus differ only in where the graph and the table come from.
The answer is a \emph{certificate}: what each arm's removal leaves undone; each object used or rejected, with its measurement; the jobs that cannot all have a doer when no team exists; and the near cells counted as reached.

\paragraph{Stand-in objects.}
Only when the search finds no team does \Scissors look for more objects: the prover, still blind, is told which jobs lack a doer and asked what on the table could take them over.
It may add objects, each with its bounds and placing job, but may change no other job; the new objects are measured and the search reruns.

\paragraph{Review with unequal burdens.}
The graph is the one part not measured, so it is reviewed.
Code flips each element (drops a job, takes it out of its group, moves a site, drops a bound) and marks those whose flip changes the answer. Two reviewer sessions, blind to each other, read the graph, the certificate and the marks; one looks for what was written easier than the task, the other for what was written harder.
Each edit must cite a tool call of its own session; code applies it and redraws the certificate.
An edit that changes nothing is dropped, one after which the task needs more arms or has no team is kept, and any other must quote the words of the goal that allow it and pass an \emph{arbiter} session.
An episode takes about four sessions (\cref{tab:main}): prover, two reviewers and, when needed, stand-in pass and arbiter.

\section{Experiments}
\label{sec:experiments}

\subsection{Experimental setup}
\label{sec:setup}

\paragraph{Methods and backbones.}
We compare \Scissors with LLM + solver, SMART-LLM-style~\cite{kannan2024smartllm}, RoCo-style~\cite{mandi2024roco} and HMAS-2-style~\cite{chen2024hmas} (\cref{sec:baselines}) on Qwen3.8-Max and Claude Sonnet 5.5, both at high reasoning effort.
Oracle-team (\cref{sec:refs}) is right by construction and left out of \cref{tab:main}; with Oracle-script it bounds task success in \cref{tab:exec}.

\paragraph{Runs.}
On each backbone every method answers each of the 480 scenes in the one-shot setting (\cref{sec:protocol}).
The answers on the 312 feasible episodes are carried out once by the shared executor (\cref{sec:executor}), with at most three time-limited Isaac runs; a failed run does not go back to the method.

\paragraph{Cost and statistics.}
The cost of an episode is the arms in the team it names (mean over the feasible episodes with a team named), its reasoning tokens (mean) and the model sessions that choose its team.
Differences are paired by scene, with 95\% bootstrap intervals over the 96 task-scenes, and a method is ahead only where the interval excludes 0.

\subsection{Main results}
\label{sec:results}

\paragraph{\Scissors names the right team most often on both backbones.}
\label{sec:overall}
It is right in 85.4\% and 91.2\% of episodes with Qwen3.8-Max and Claude Sonnet 5.5, against 75.0\% and 81.7\% for the next method (HMAS-2-style, SMART-LLM-style), and first on both task families, on which its scores differ by at most 4.2 points.
The stronger backbone raises every method, by 3.8 (HMAS-2-style) to 15.2 points (RoCo-style), but does not close the gap: \Scissors with Qwen3.8-Max is right more often than any other method with
Claude Sonnet 5.5.
The gap is widest at the task level, the headline of \cref{sec:metrics}: \Scissors gets all five configurations of a task right in 43.8\% and 59.4\% of the task-scenes, the other methods in 2.1--17.7\% and 18.8--32.3\%, so that with Qwen3.8-Max all of them fall below the best fixed rule (26.0\%, \cref{sec:gold}), and with Claude Sonnet 5.5 all but SMART-LLM-style and HMAS-2-style.
Paired by scene, \Scissors is ahead of every method on both backbones, by 9.5 to 24.4 points over all episodes and 26.1 to 41.7 at the task level, with no 95\% interval including 0.

\paragraph{A better translation does not tell what an arm can do.}
\label{sec:translate}
LLM + solver runs the search of \Scissors on the model's one-pass translation, in which the model also writes what each arm can do (\cref{sec:method}); \Scissors leads it by 19.2 and 13.1 points.
The stronger backbone closes most of the gap on \ver{shortfall}, from 33.4 to 5.2 points, where the key arm is simply absent, but not on \ver{swapped} (33.3 and 27.0 points), where what a new arm reaches has to be known.
A better translation thus does not supply \textbf{what the arms can do}, which is what \Scissors measures in code.

\paragraph{A spare arm is mostly left idle, and a stronger model does not help.}
\label{sec:surplus}
Every method is right in at least 76.0\% of the \ver{surplus} episodes, and \Scissors in 96.9\% on both backbones.
With the stronger backbone three of the other methods fall, by 4.1 to 9.4 points, and only SMART-LLM-style rises (80.2\% to 82.3\%).
Of the 151 wrong \ver{surplus} answers on both backbones, 30.5\% staff the spare arm where it is not needed, \emph{staffing by presence}; as many name too few arms, and 21.2\% refuse the feasible task.

\begin{table}[t]
  \caption{\textbf{Task success with right and wrong teams.} Success (\%) on the 312 feasible episodes. Team: right team named; SR\,\textbar\,right, SR\,\textbar\,wrong: success with a right or a wrong team.}
  \label{tab:exec}
  \centering
  \footnotesize
  \setlength{\tabcolsep}{1pt}
  \newcommand{\hd}[1]{{\fontsize{6.5}{7.5}\selectfont\makecell{#1}}}
  \begin{NiceTabular*}{\linewidth}{@{\extracolsep{\fill}}l*{6}{c}@{}}[cell-space-limits=0.8pt]
    \toprule
    \Block[l]{2-1}{Method} & \Block{1-3}{Qwen3.8-Max} & & & \Block{1-3}{Claude Sonnet 5.5} & & \\
    \cmidrule(lr){2-4} \cmidrule(lr){5-7}
    & \hd{Team} & \hd{SR\,\textbar\\right} & \hd{SR\,\textbar\\wrong} & \hd{Team} & \hd{SR\,\textbar\\right} & \hd{SR\,\textbar\\wrong} \\
    \midrule
    Oracle-script   & 100  & 100  & \na  & 100  & 100  & \na  \\
    Oracle-team     & 100  & 58.7 & \na  & 100  & 60.3 & \na  \\
    \midrule
    LLM + solver    & 72.4 & 20.8 & 1.2 & 75.3 & 25.1 & 1.3 \\
    SMART-LLM-style & 72.4 & 25.2 & 4.7 & 79.2 & 27.5 & 7.7 \\
    RoCo-style      & 67.9 & 18.9 & 1.0 & 71.5 & 23.3 & 0.0 \\
    HMAS-2-style    & 73.7 & 23.0 & 6.1 & 76.0 & 27.4 & 9.3 \\
    \midrule
    \textbf{\Scissors (ours)} & \textbf{83.7} & \textbf{55.9} & 0.0 & \textbf{87.8} & \textbf{58.8} & 0.0 \\
    \bottomrule
  \end{NiceTabular*}
\end{table}

\paragraph{A stronger backbone teaches every method to say no.}
\label{sec:shortfall}
On \ver{shortfall} the right answer is infeasible.
With Qwen3.8-Max \Scissors gives it in 93.8\% of the episodes and LLM + solver and RoCo-style in only 60.4\% and 52.1\%; with Claude Sonnet 5.5 every method gives it in at least 91.7\%, LLM + solver and RoCo-style gaining 32.3 and 39.6 points, the largest gains in \cref{tab:main}.
Since ``always infeasible'' is right on every \ver{shortfall}, refusing a feasible task is its cost: \Scissors refuses 11 and 12 of its 312 feasible episodes, the other methods 20--31 and 15--27.

\paragraph{A swapped arm separates the methods most.}
\label{sec:swapped}
\Scissors is right in 83.3\% and 95.8\% of the \ver{swapped} episodes, the other methods in 46.9--64.6\% and 68.8--75.0\%; its lead over the next method, 18.7 and 20.8 points, is its largest in any configuration.
The other methods do not read the SO-101 as infeasible: pooled, they are right about as often in the swaps that bring one in, all of which make the task infeasible (51.0\% and 71.4\%), as in the other swaps (56.2\% and 72.4\%).
Where a Franka replacing a PiPER lets two arms suffice (Rea-02, Rea-04), so that keeping the old team, \emph{roles by place}, is wrong, they are right in 45.8--62.5\% and 58.3--75.0\%, and \Scissors in 87.5\% and 91.7\%.

\paragraph{Whether a passive object can stand in for an arm remains open.}
\label{sec:borrow}
\ver{borrow} is the weakest configuration of every method on both backbones, except LLM + solver with Qwen3.8-Max, which does 1.0 point worse on \ver{swapped}.
\Scissors is first here too, at 59.4\% and 66.7\% against at most 51.0\% and 59.4\%, yet misses a third of these episodes even with the stronger backbone.
Choosing by size, \emph{substitution by appearance}, does not explain the errors: no method does more than 5.0 points better where the prop is the larger object (67 of the 96 scenes).
What remains hard is whether the object can do the job at all.
Of the 450 wrong \ver{borrow} answers, 28.4\% use the distractor and 31.8\% call the task infeasible; the other 39.8\% use the prop but name too few arms, as if it did more than the missing arm's job.

\paragraph{\Scissors uses fewer tokens than every adapted method.}
\label{sec:cost}
It uses 421.5\,k and 452.5\,k tokens per episode, a third and a quarter fewer than the cheapest adapted method, SMART-LLM-style (640.6\,k and 610.0\,k), and at most 0.6 times those of HMAS-2-style and RoCo-style. Its 3.8 and 3.9 model sessions per episode lie between SMART-LLM-style (3.0) and HMAS-2-style (7.8 and 6.1).
Only LLM + solver, at one session, costs less (327.8\,k and 282.2\,k tokens), and it is right 19.2 and 13.1 points less often.

\paragraph{The lead holds on held-out seeds.}
\label{sec:heldout}
No method was developed on seeds 1--5: on their 400 scenes, across all 40 versions, \Scissors leads the next method by 11.4 and 9.4 points.
It is also first on every task, by 1.7 to 16.7 points, and at both levels.

\paragraph{Errors by kind.}
\label{sec:errors}
Of \Scissors' 70 and 42 wrong answers, 22 and 14 name too few arms, 19 and 4 name a team for an infeasible task, 11 and 12 refuse a feasible one, 16 and 9 use the distractor instead of the prop, and 2 and 3 name the right number of arms but not the right ones; none names an arm too many, as the others do in 4 to 18 answers.

\subsection{From the right team to task success}
\label{sec:exec}

\paragraph{Right teams carry different solutions.}
The executor carries out a method's whole answer (\cref{sec:executor}), so SR and MSS-SR score its solution as well as its team.
\Scissors' steps, which arm does each job in what order, are read off its certificate, where each job's doer is a measured cell; in the other methods the model writes them.
On all 312 feasible episodes, \Scissors succeeds with a right team (MSS-SR) in 46.8\% and 51.6\%, more than twice the 12.8--18.3\% and 16.7--21.8\% of the other methods; on the episodes where it and another method both name a right team, it succeeds in 51.5--55.4\% and 56.9--58.2\%, the other method in 17.9--24.0\% and 24.0--27.4\%. Oracle-team, handed the gold team, reaches 58.7\% and 60.3\% (\cref{tab:exec}).
With a right team, \Scissors' own episodes succeed in 55.9\% and 58.8\%, within 3 points of that ceiling.

\paragraph{Success alone rewards an arm too many.}
SR exceeds MSS-SR in seven of the ten rows of \cref{tab:main}: in each of these rows 1 to 7 episodes succeed with a team that is not the right one (\cref{tab:exec}, SR\,\textbar\,wrong), every one of them with an arm too many; \Scissors names no such team.
A success shows that a team suffices, not that it is needed, which is why \Bench scores task success as MSS-SR.

\subsection{Ablation study}
\label{sec:ablation}

\begin{table}[t]
  \caption{\textbf{Ablation on \Scissors' parts.} Accuracy (\%) on 80 Qwen3.8-Max episodes at the easy level, seeds 0 and 4; sessions per episode. Meas.: measured table; Rev.: reviewers; Arb.: arbiter.}
  \label{tab:ablation}
  \centering
  \scriptsize
  \setlength{\tabcolsep}{0.9pt}
  \newcommand{\hd}[1]{{\fontsize{6}{7}\selectfont\makecell{#1}}}
  \begin{NiceTabular*}{\linewidth}{@{\extracolsep{\fill}}c*{7}{c}*{6}{c}@{}}[cell-space-limits=0.6pt]
    \CodeBefore
      \rowcolor{slGreen!10}{7}
      \rowcolor{black!4}{12,13}
    \Body
    \toprule
    & \hd{Meas.} & \hd{Blind} & \hd{Stand-\\in} & \hd{Flip} & \hd{Rev.} & \hd{Arb.} & \hd{Near} & \hd{\ver{swapped}} & \hd{\ver{borrow}} & \hd{Feasible} & \hd{Infeasible} & \hd{All} & \hd{Sessions} \\
    \midrule
    (a) &      &      &      &      &   &      &      & 62.5 & 37.5 & 71.2 & 53.6 & 65.0 & 1.0 \\
    (b) & \yes &      &      &      &   &      & \yes & 75.0 & 25.0 & 61.5 & 78.6 & 67.5 & 1.0 \\
    (c) & \yes & \yes &      &      &   &      & \yes & 81.2 & 6.2 & 59.6 & \textbf{89.3} & 70.0 & 1.0 \\
    (d) & \yes & \yes & \yes &      &   &      & \yes & 81.2 & 43.8 & 71.2 & \textbf{89.3} & 77.5 & 1.6 \\
    (e) & \yes & \yes & \yes & \yes & 2 &      & \yes & 81.2 & 56.2 & 82.7 & 82.1 & 82.5 & 3.8 \\
    (f) & \yes & \yes & \yes & \yes & 2 & \yes & \yes & 75.0 & \textbf{68.8} & \textbf{88.5} & 78.6 & \textbf{85.0} & 3.9 \\
    \midrule
    (g) & \yes &      & \yes & \yes & 2 & \yes & \yes & 75.0 & 62.5 & 84.6 & 78.6 & 82.5 & 3.8 \\
    (h) & \yes & \yes & \yes &      & 2 & \yes & \yes & 81.2 & 50.0 & 76.9 & \textbf{89.3} & 81.2 & 3.6 \\
    (i) & \yes & \yes & \yes & \yes & 1 & \yes & \yes & 75.0 & 56.2 & 78.8 & 85.7 & 81.2 & 2.8 \\
    (j) & \yes & \yes & \yes & \yes & 2 & \yes &      & \textbf{87.5} & 56.2 & 82.7 & \textbf{89.3} & \textbf{85.0} & 3.8 \\
    \midrule
    (k) & \Block[l]{1-7}{Gold graph + measured table} & & & & & & & 93.8 & 93.8 & 98.1 & 96.4 & 97.5 & \na \\
    (l) & \Block[l]{1-7}{Own graph + gold table} & & & & & & & 87.5 & 62.5 & 86.5 & 89.3 & 87.5 & 3.8 \\
    (m) & \Block[l]{1-7}{SMART-LLM-style + table} & & & & & & & 56.2 & 43.8 & 71.2 & 78.6 & 73.8 & 4.8 \\
    (n) & \Block[l]{1-7}{HMAS-2-style + table} & & & & & & & 68.8 & 43.8 & 73.1 & 78.6 & 75.0 & 9.6 \\
    \bottomrule
  \end{NiceTabular*}
\end{table}

\Cref{tab:ablation} adds the parts of \Scissors to LLM + solver one at a time (a--f), removes each (g--j), gives it gold inputs (k,~l) and hands its table to two adapted methods (m,~n).

\paragraph{Caution first, then recovery.}
The first two parts make \Scissors cautious.
Measuring the capability table in code (b) and hiding the arms from the prover (c) together raise accuracy on infeasible episodes from 53.6\% to 89.3\%, but lower it on feasible ones from 71.2\% to 59.6\%, most of all on \ver{borrow}, where hands alone fall short.
The stand-in pass (d) and the review (e,~f) win these episodes back: feasible accuracy rises to 88.5\%, while infeasible accuracy gives back only part of its gain (78.6\%).
Overall accuracy therefore rises at every step, from 65.0\% to 85.0\%, although no single column does.
Taking out the blind prover, the flip marks, a reviewer or the arbiter lowers it by 2 or 3 of the 80 episodes (g--i,~e); counting near cells as unreachable does not (j).
The parts thus trade \textbf{caution against coverage}: the later ones recover what the earlier ones cost.

\paragraph{Most remaining errors lie in the job graph.}
Given the gold job graph with a measured table (k), \Scissors is right in 97.5\% of the episodes; given its own graph with the gold capabilities (l), in 87.5\%.
Of its 12 errors, the gold graph thus removes 10 and the gold table 2.
Most errors come from \textbf{what the model writes}, the jobs, their groups and a stand-in's bounds, not from what the tools measure.

\paragraph{The procedure, not the table alone, closes the gap.}
Every method has the same tools, but only \Scissors calls them for every arm at every site.
Handed \Scissors' measured table, at 1.75 more sessions per episode, SMART-LLM-style and HMAS-2-style reach 73.8\% and 75.0\% (m,~n), against 85.0\% for \Scissors.
The gain thus lies not in the measurements, which any method could adopt, but in \textbf{the procedure that turns them into an answer}.

\section{Conclusion}
\label{sec:conclusion}

Before arms can cooperate, an agent must decide how many a task needs, and that decision must rest on measurement.
\Bench turns it into a test: eight tasks on heterogeneous arms, five configurations each that change the minimal feasible team under one goal, and audited gold answers.
\Scissors answers it from measured evidence: a solver choosing over a measured table names the right team in 85.4\% and 91.2\% of episodes on two backbones, against at most 75.0\% and 81.7\% for the other methods.
Yet \Scissors gets all five configurations right in at most 59.4\% of the task-scenes and misses at least a third of the \ver{borrow} episodes: what can stand in for an arm is the open problem.

{
    \small
    \bibliographystyle{ieeenat_fullname}
    \bibliography{A-reference}
}

\clearpage
\setcounter{dbltopnumber}{3}\renewcommand{\dbltopfraction}{0.85}\renewcommand{\dblfloatpagefraction}{0.6}
\setcounter{topnumber}{3}\renewcommand{\topfraction}{0.85}\renewcommand{\floatpagefraction}{0.6}\renewcommand{\textfraction}{0.1}
\makeatletter\setlength{\@fptop}{0pt}\setlength{\@fpsep}{\floatsep}\setlength{\@fpbot}{0pt plus 1fil}\setlength{\@dblfptop}{0pt}\setlength{\@dblfpsep}{\dblfloatsep}\setlength{\@dblfpbot}{0pt plus 1fil}\makeatother

\appendix
\renewcommand{\thesection}{\Alph{section}}

\section{Appendix}
\label{sec:appendix}

\subsection{Task cards}
\label{sec:supp-tasks}

\begin{figure*}[t]
  \centering
  \newcommand{\fr}[2]{\begin{minipage}[t]{0.1625\textwidth}\centering\includegraphics[width=\linewidth]{figures/supp/#1}\\[-1pt]{\scriptsize\strut #2}\end{minipage}}
  \newcommand{\ttl}[3]{{\small\textbf{#1} #2\hfill\textcolor{slGray}{#3}}}
  \ttl{Con-01}{spring doors}{kitchen}\\[1pt]
  \fr{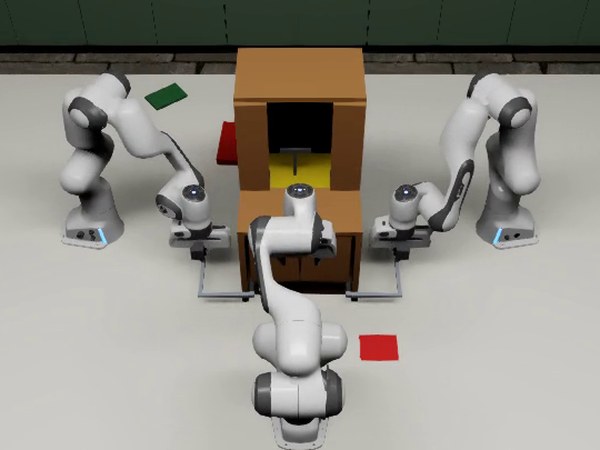}{(1) \W, \E take the knobs}\hfill
  \fr{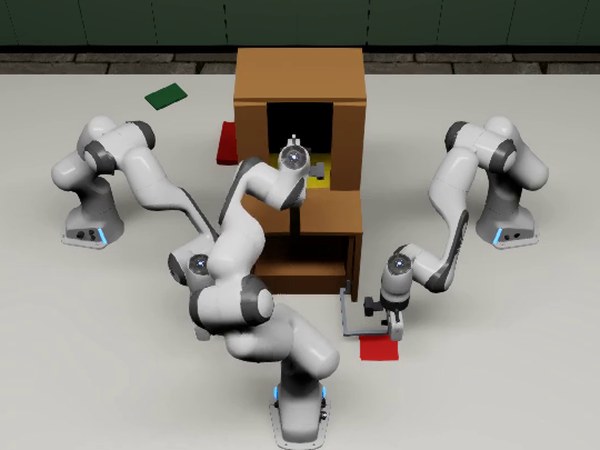}{(2) and hold the doors open}\hfill
  \fr{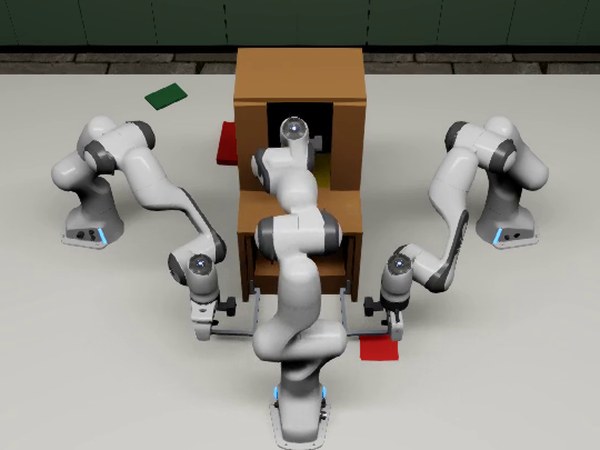}{(3) \Ss takes the box}\hfill
  \fr{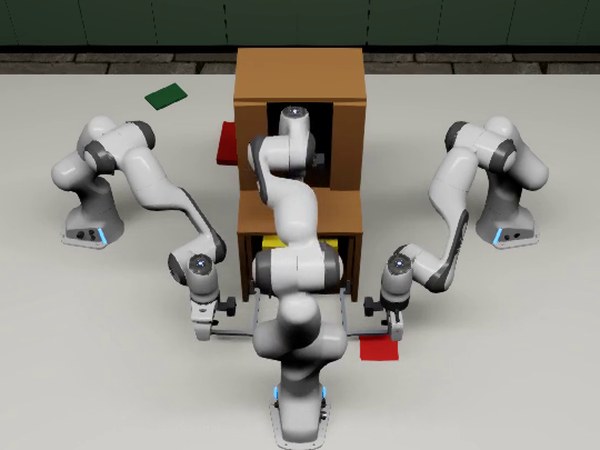}{(4) draws it out}\hfill
  \fr{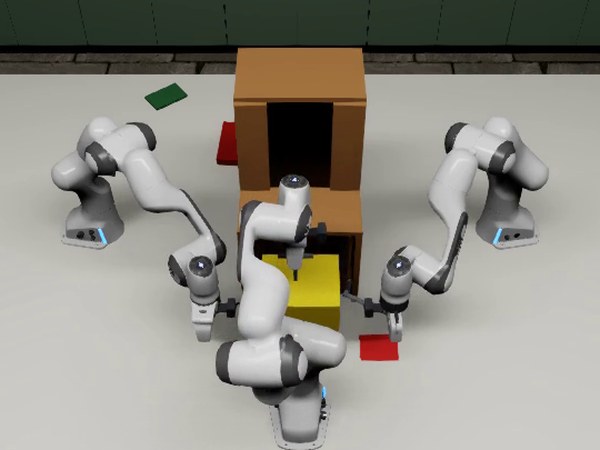}{(5) and out; the doors shut}\hfill
  \fr{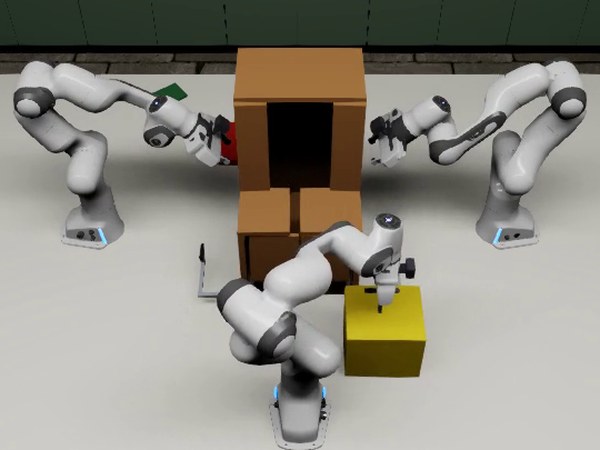}{(6) the box on the mat}\\[14pt]
  \ttl{Con-02}{hinged leaves}{workshop}\\[1pt]
  \fr{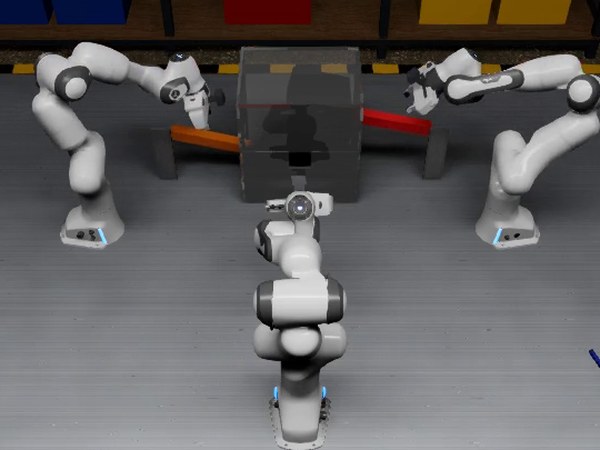}{(1) \W, \E reach the leaves}\hfill
  \fr{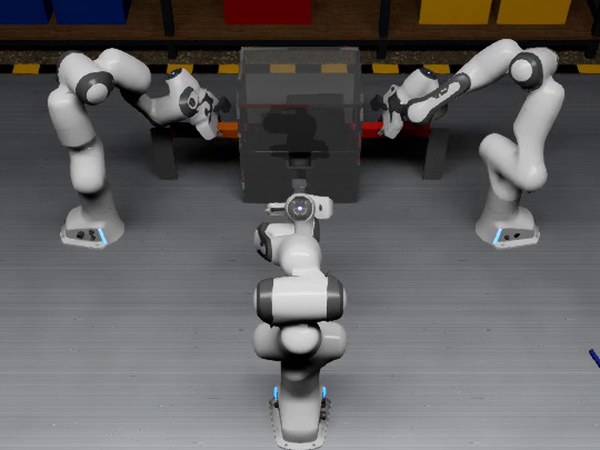}{(2) and turn them level}\hfill
  \fr{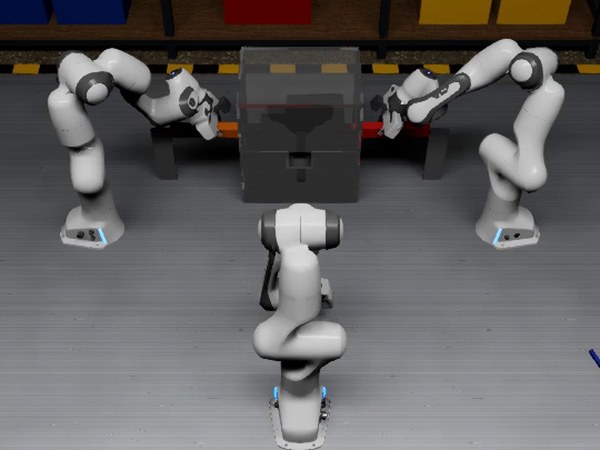}{(3) \Ss takes the pin}\hfill
  \fr{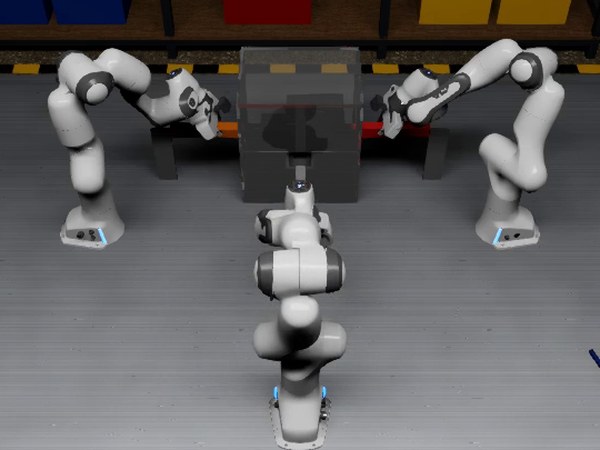}{(4) and lines it up}\hfill
  \fr{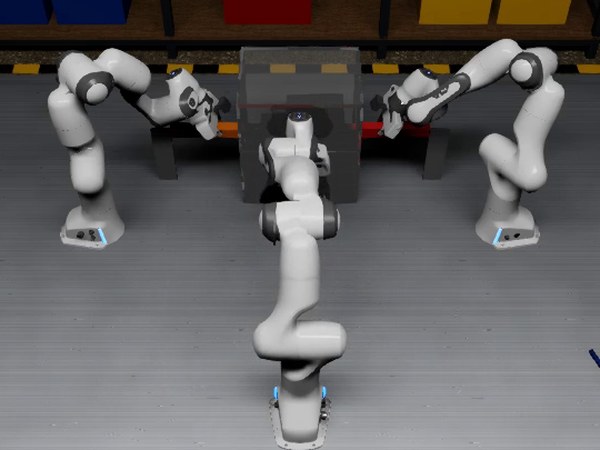}{(5) slides it through the slots}\hfill
  \fr{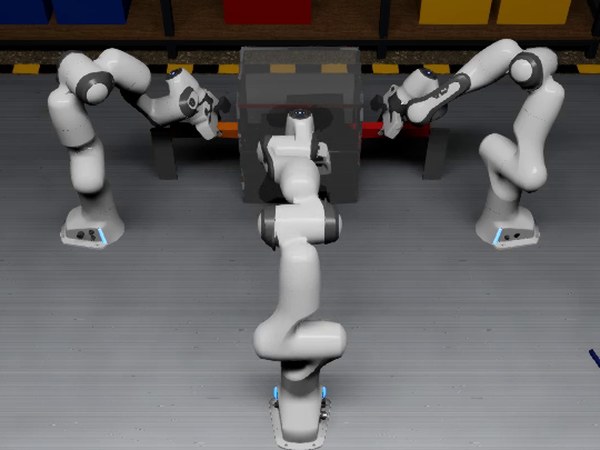}{(6) pinned and level}\\[14pt]
  \ttl{Con-03}{spring lid}{home}\\[1pt]
  \fr{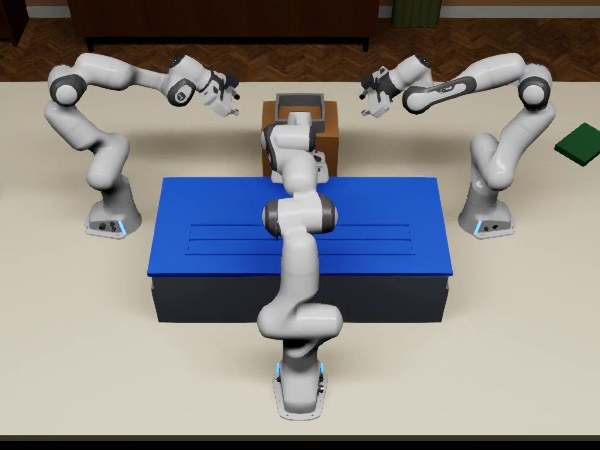}{(1) \Ss takes the lid's tab}\hfill
  \fr{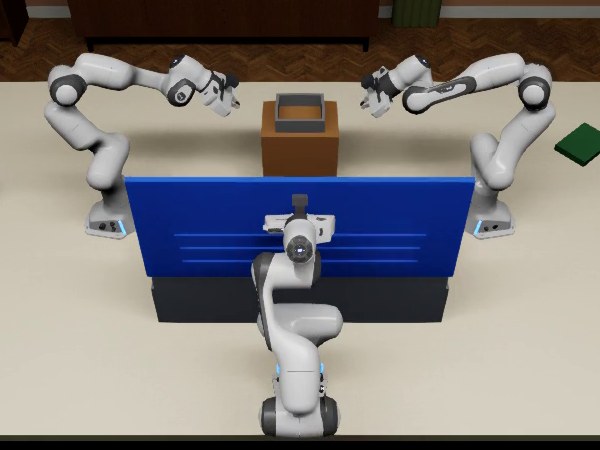}{(2) and holds the lid open}\hfill
  \fr{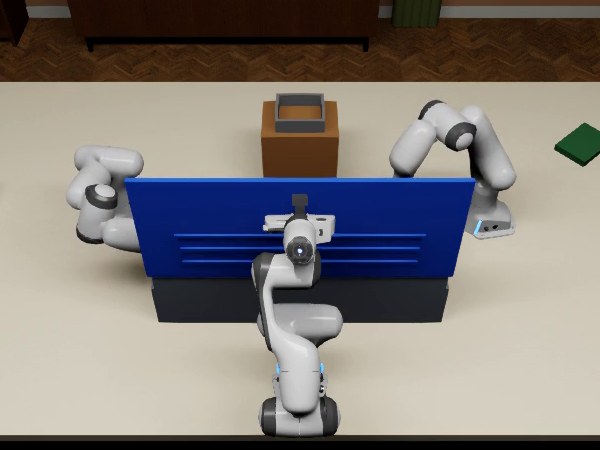}{(3) \W, \E press the ball}\hfill
  \fr{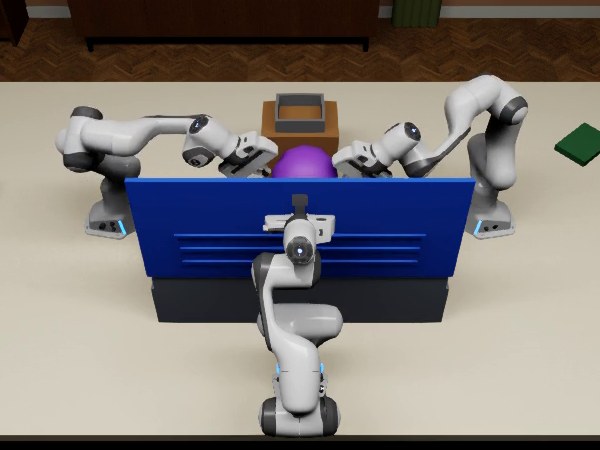}{(4) and lift it out}\hfill
  \fr{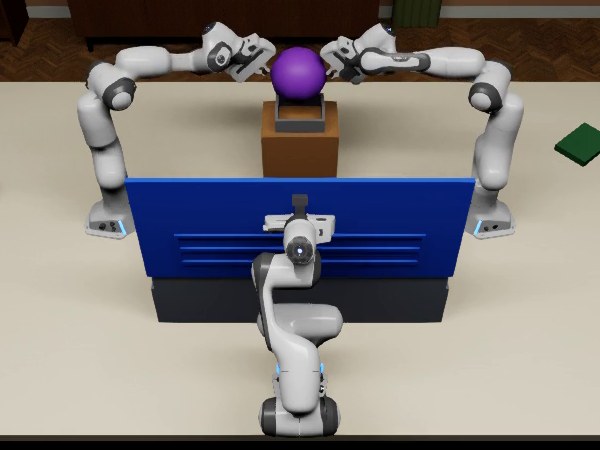}{(5) carry it to the cradle}\hfill
  \fr{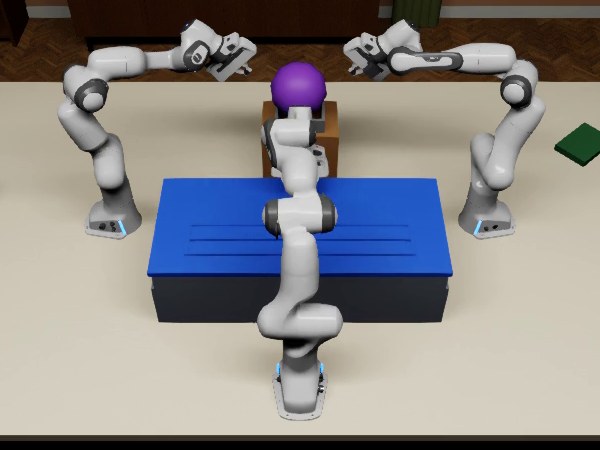}{(6) set in; \Ss lets the lid down}\\[14pt]
  \ttl{Con-04}{pressure test}{kitchen}\\[1pt]
  \fr{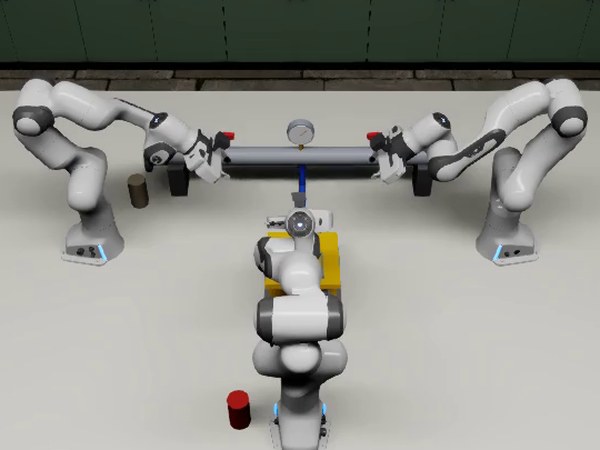}{(1) the start}\hfill
  \fr{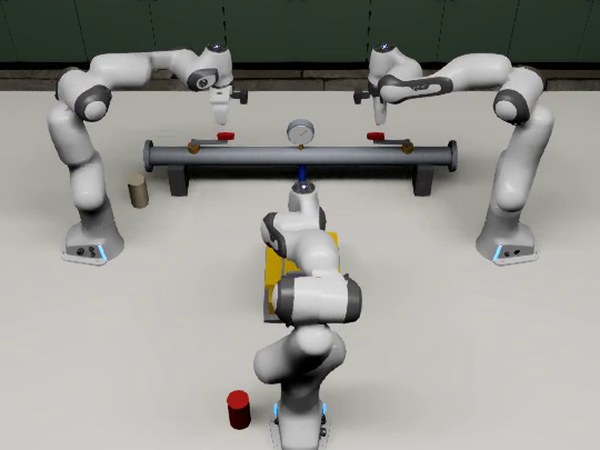}{(2) the hands come over}\hfill
  \fr{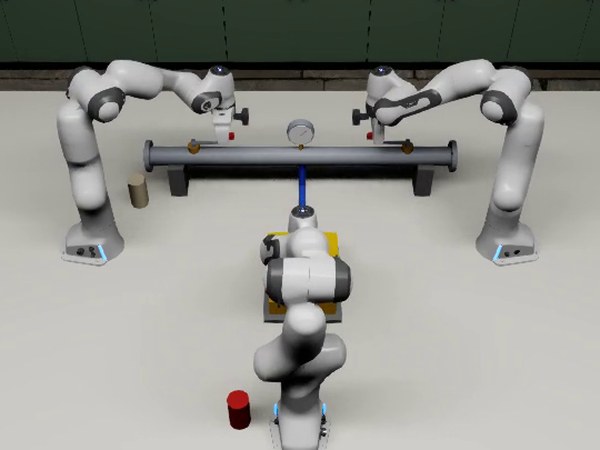}{(3) \W, \E, \Ss take them}\hfill
  \fr{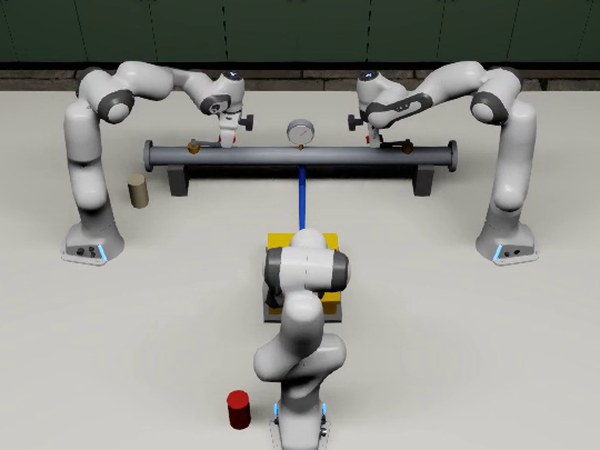}{(4) pressed: pressure held}\hfill
  \fr{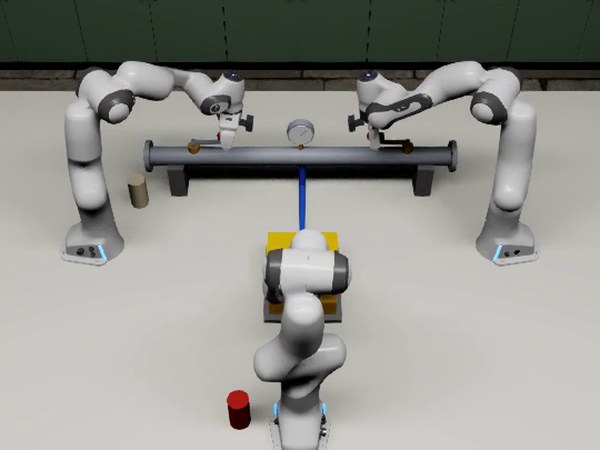}{(5) they let go}\hfill
  \fr{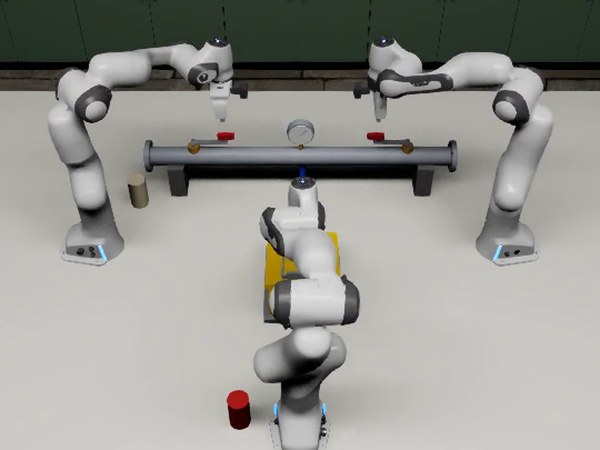}{(6) the springs lift them}
  \caption{\textbf{The \Con references in Isaac.} Each task's scripted reference at the easy level in one varied scene, as in its task card; the room, the colors of the task's objects and the everyday objects on the table follow the seed.}
  \label{fig:supp-ref-con}
\end{figure*}

\begin{figure*}[t]
  \centering
  \newcommand{\fr}[2]{\begin{minipage}[t]{0.1625\textwidth}\centering\includegraphics[width=\linewidth]{figures/supp/#1}\\[-1pt]{\scriptsize\strut #2}\end{minipage}}
  \newcommand{\ttl}[3]{{\small\textbf{#1} #2\hfill\textcolor{slGray}{#3}}}
  \ttl{Rea-01}{cross-zone relay}{workshop}\\[1pt]
  \fr{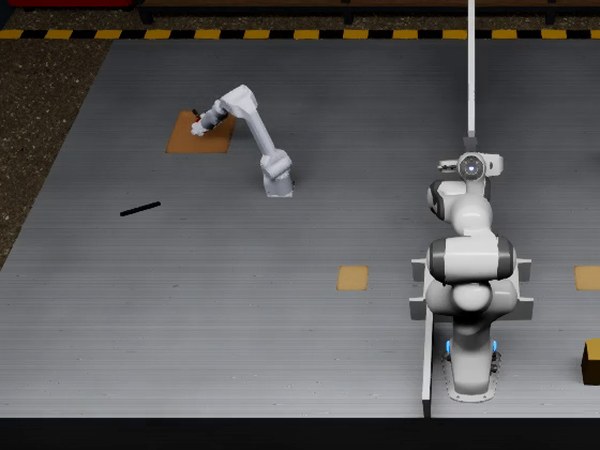}{(1) \W takes the apple}\hfill
  \fr{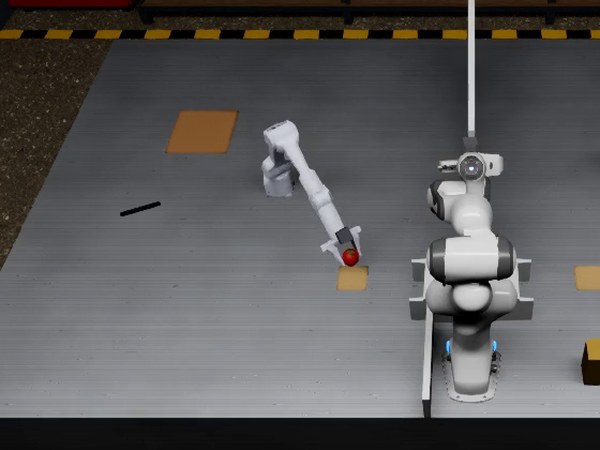}{(2) and sets it on pad A}\hfill
  \fr{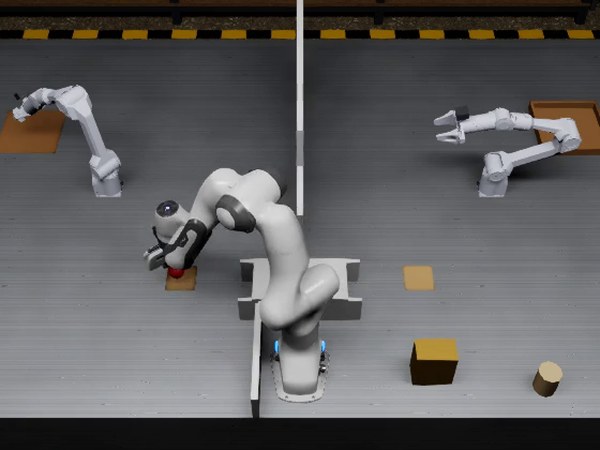}{(3) \Ss takes it}\hfill
  \fr{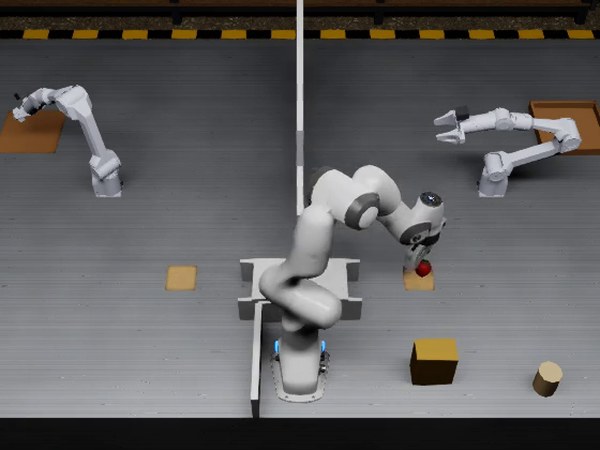}{(4) and sets it on pad B}\hfill
  \fr{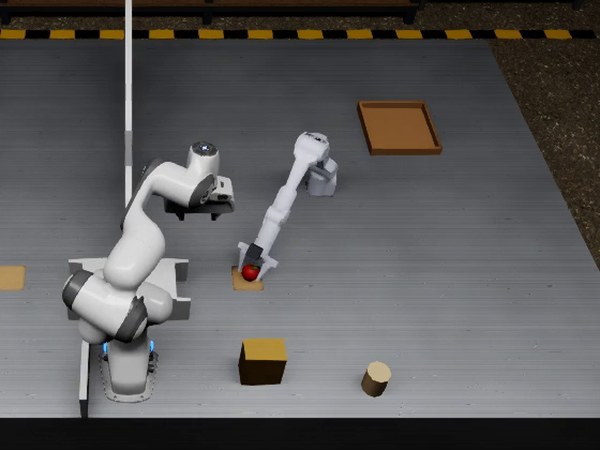}{(5) \E takes it}\hfill
  \fr{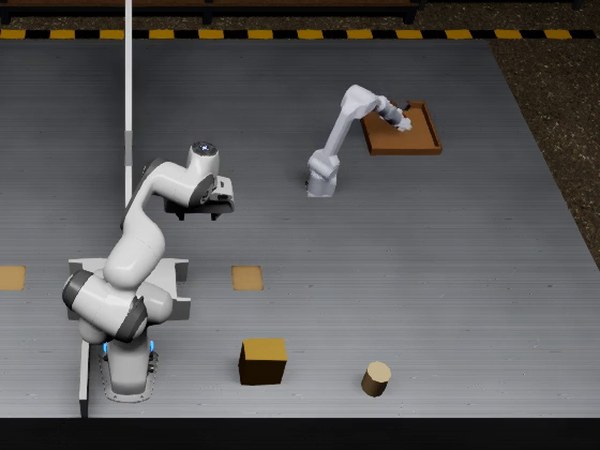}{(6) and puts it in the crate}\\[14pt]
  \ttl{Rea-02}{tomato flume}{home}\\[1pt]
  \fr{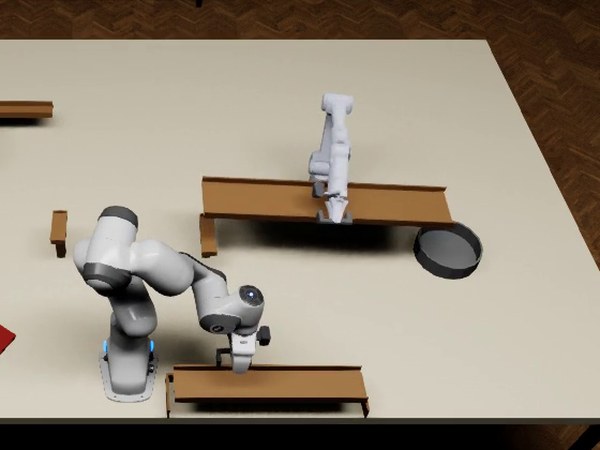}{(1) \E lays the lowest channel}\hfill
  \fr{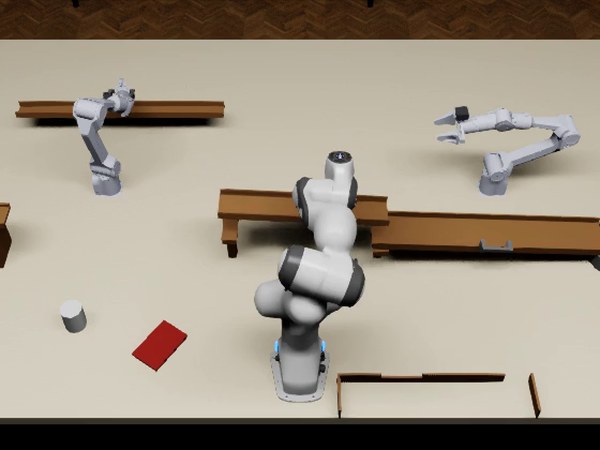}{(2) \Ss the middle one}\hfill
  \fr{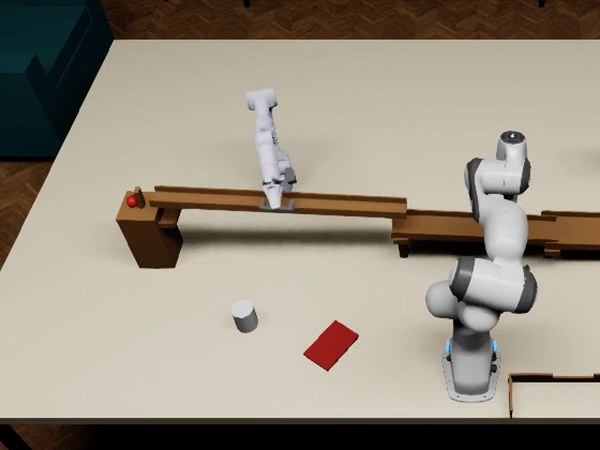}{(3) \W the upper one}\hfill
  \fr{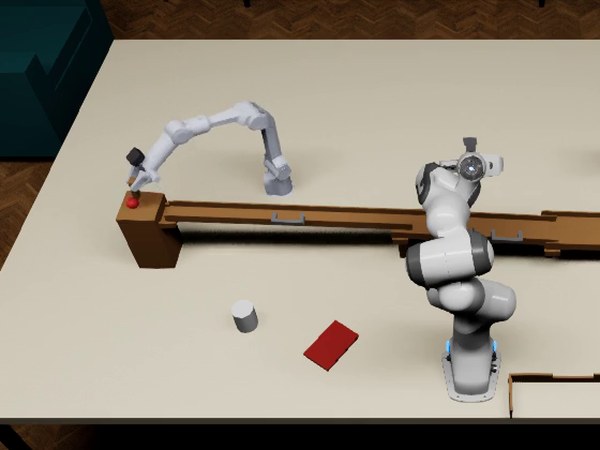}{(4) \W lifts the stop}\hfill
  \fr{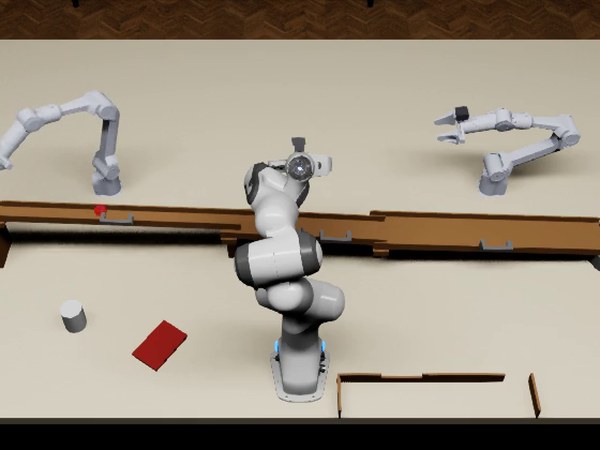}{(5) the tomato rolls down}\hfill
  \fr{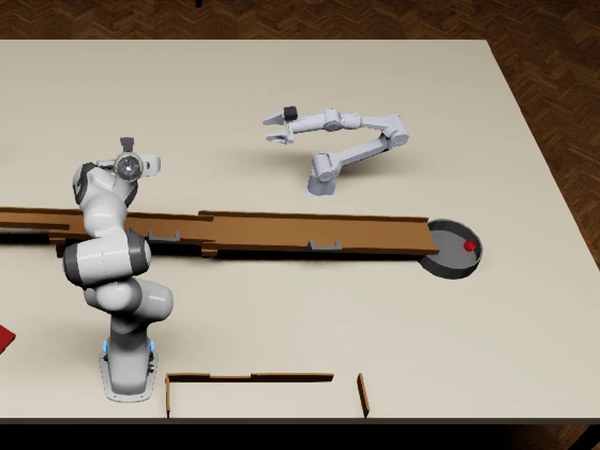}{(6) into the pot}\\[14pt]
  \ttl{Rea-03}{domino run}{kitchen}\\[1pt]
  \fr{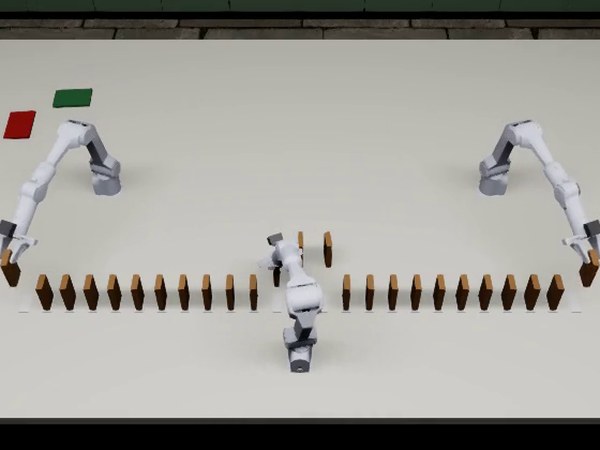}{(1) each carries a domino}\hfill
  \fr{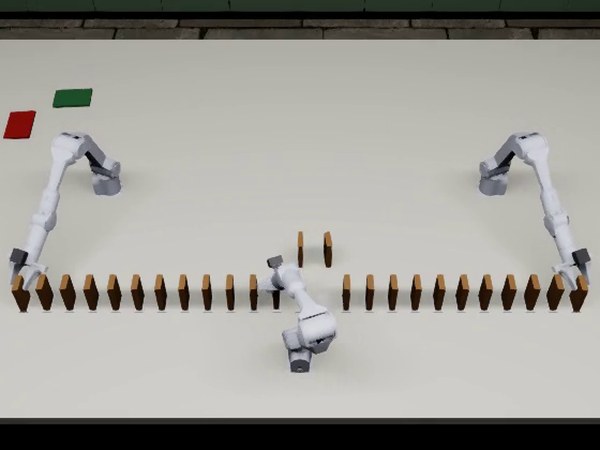}{(2) the end places filled}\hfill
  \fr{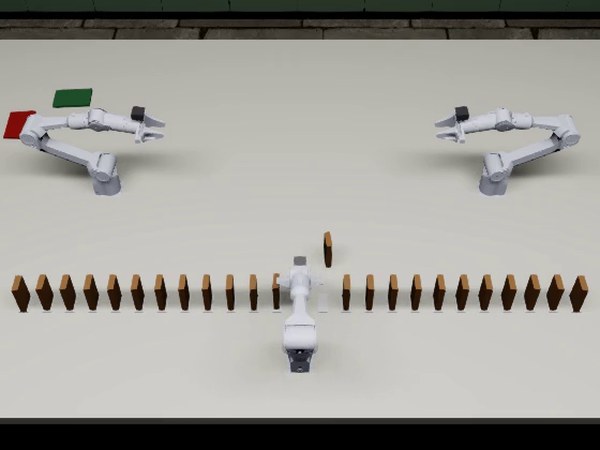}{(3) \Ss fills the middle gap}\hfill
  \fr{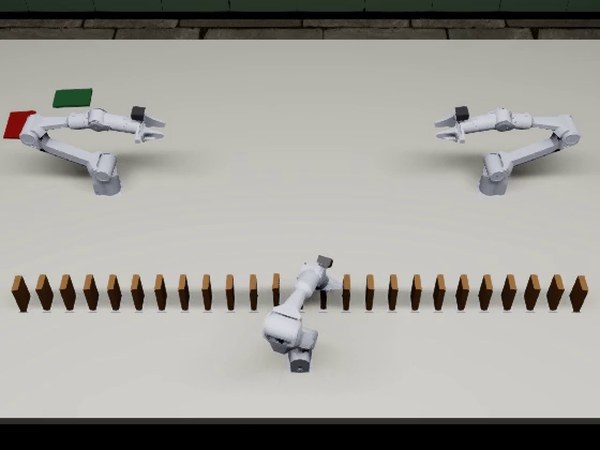}{(4) the line is whole}\hfill
  \fr{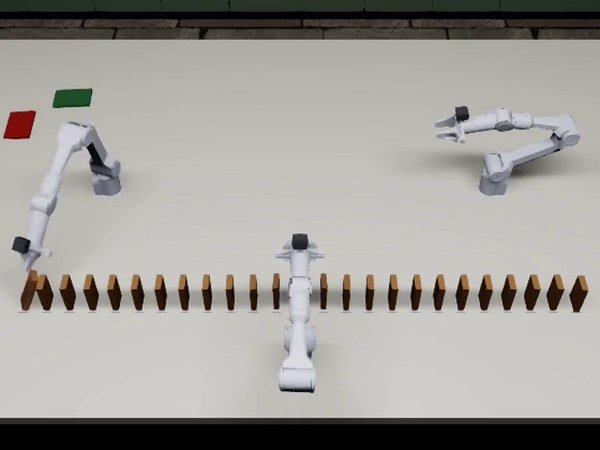}{(5) \W tips the first domino}\hfill
  \fr{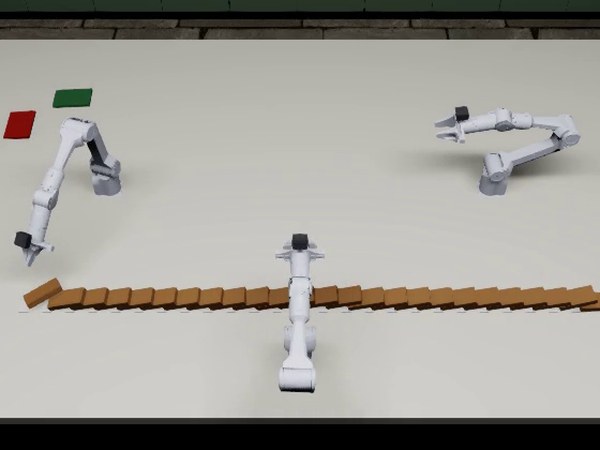}{(6) the whole line is down}\\[14pt]
  \ttl{Rea-04}{wipe the spills}{workshop}\\[1pt]
  \fr{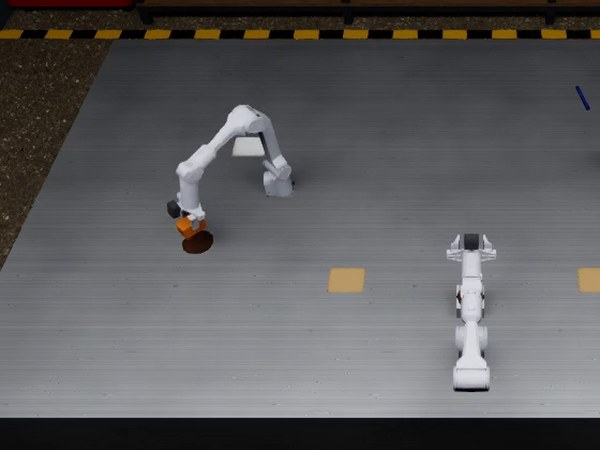}{(1) \W wipes its spill}\hfill
  \fr{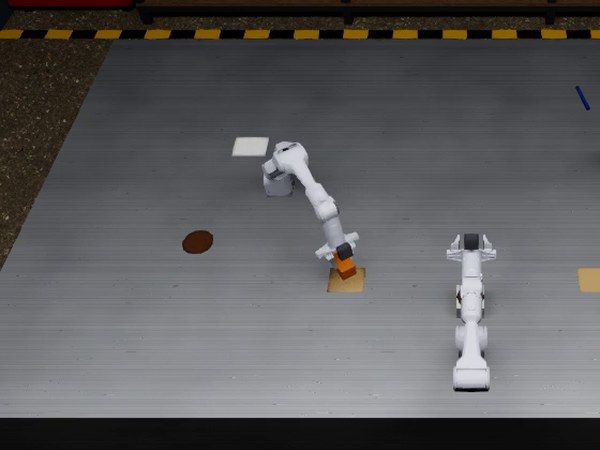}{(2) and puts it on pad A}\hfill
  \fr{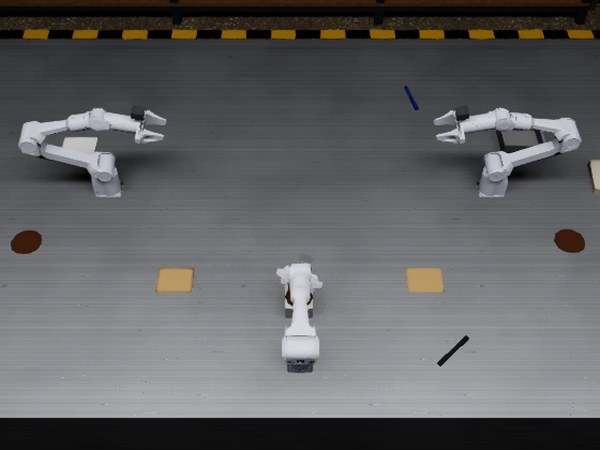}{(3) \Ss wipes the middle spill}\hfill
  \fr{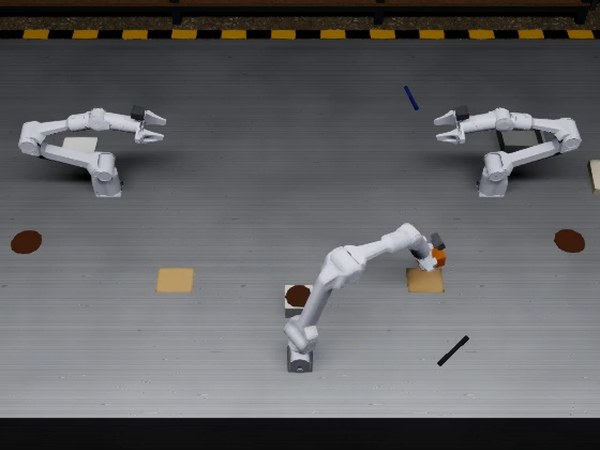}{(4) and sets it on pad B}\hfill
  \fr{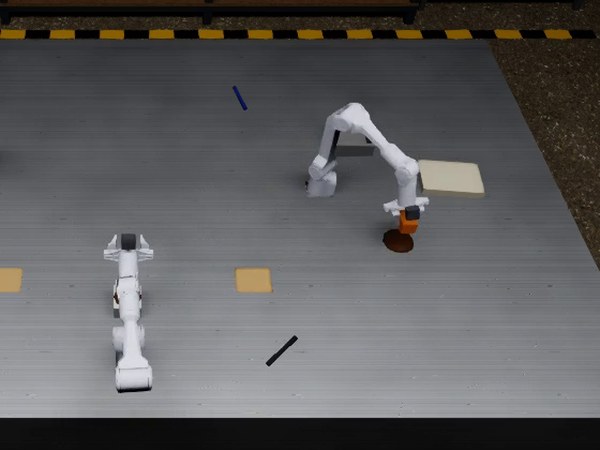}{(5) \E wipes its spill}\hfill
  \fr{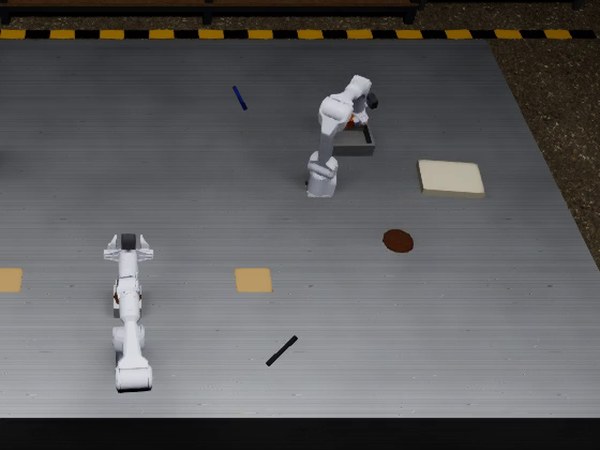}{(6) and drops it in the sink}
  \caption{\textbf{The \Rea references in Isaac.} As in \cref{fig:supp-ref-con}; each arm works only in its own part of the table, and the work passes from arm to arm. Each frame shows the part of the table where its arm works.}
  \label{fig:supp-ref-rea}
\end{figure*}

Each card gives the instruction, the arms at \W/\Ss/\E, the parameter that sets the two levels, why three arms are needed, the scripted reference with its duration in Isaac Sim at the easy / hard level (seed 0; a single figure when both levels take the same time), the success predicate, checked on the objects' states, and what the necessity audit checks.
The instruction is given as in the default scene: in the varied scenes it names the task's objects by the colors the seed gives them, and Rea-01 and Rea-04 name their level's number of fruit and wipe time.
Instructions state the goal, and three also bound the process: Rea-02's tomato must roll, not be carried, and success asks it to reach the pot within 5\,s of leaving the shelf; Rea-03's line must be whole when it is set off; Rea-04's sponge must stay on each spill for the wipe time.
The gold answers are smallest under these bounds.
\Cref{fig:supp-ref-con,fig:supp-ref-rea} show each reference in a varied scene.

\paragraph{Con-01, spring doors.}
\emph{``Take the orange box out of the cabinet and set it on the green mat, with both doors closed.''}
Three Frankas; box width 23 / 26\,cm.
A cabinet stands across the middle of the table, its doorway facing \Ss.
In front, two spring doors as high as the box open inward under a fixed ceiling; each is worked by a low lever outside it, a bar beside the side wall with a knob.
The box is wider than one door's way and deeper and taller than the slot over the ceiling, so no turning gets it out: both doors must be held open while a third hand draws the box out by its T-handle, whose post runs in a narrow slot in the ceiling.
\emph{Reference} (35.0\,s): \W and \E each swing their lever round and hold their door at 88\textdegree; \Ss draws the box out under the ceiling, takes it again from above, holds it up while the doors close, and sets it on the mat.
\emph{Success.} The box has left the cabinet and rests on the mat with both doors closed, for 1\,s and to the end of the run.
\emph{Audit.} The ceiling and the T-handle keep every link of the retriever clear of the doors (IK: at least 1.4\,cm at 81--88\textdegree).
With \W{}+\Ss alone, the east door stays shut and blocks the box.
Grasps refuse the box's 23--26\,cm face.

\paragraph{Con-02, hinged leaves.}
\emph{``Pin the blue and the orange leaf together through all their slots with the grey pin, so that they hold each other level between their posts.''}
Three Frankas; slot clearance 6 / 3\,mm.
Two leaves hinged on posts west and east meet in the air as a clevis inside a glass guard; their slots line up only while both are held level, since, let go, the west leaf hangs down and the east one springs up.
Each leaf enters the guard through a slot in its side wall, and only the arm at its end reaches its bar; the pin goes in through a small opening in the south wall.
\emph{Reference} (25.2\,s): \W and \E turn their leaves level and hold them; \Ss slides the pin through the opening and the three slots; pinned, the leaves hold each other level.
\emph{Success.} The pin runs through all three slots and both leaves are level within 3\textdegree, for 1\,s.
\emph{Audit.} Hinges and guard leave no two-arm route, such as pinning the parts on the table and lifting them (IK scans: no other arm reaches a leaf; one arm holding both comes no nearer than 17\,cm).
Geometry rules out setting the pin, a flat key, in one leaf's slot and using it as a lever on the other, so this route is excluded without a run.

\paragraph{Con-03, spring lid and basketball.}
\emph{``Take the orange ball out of the box with the spring lid and set it in the cradle north of the box.''}
Three Frankas; lid spring 0.3 / 0.6\,N\,m/rad.
A long box with slick, lipped inner walls holds a 22\,cm ball, wider than any hand opens (8\,cm); its lid, hinged along the south edge on a spring, has no angle at which it stays open.
One hand holds the lid open while two others take the ball between them and set it in a cradle 54\,cm north.
\emph{Reference} (28.4 / 30.9\,s): \Ss opens the lid upright by its tab; \W and \E press their closed fingertips into the ball's sides, lift and carry it to the cradle; \Ss lets the lid down.
\emph{Success.} The ball sits in the cradle for 1\,s and to the end of the run.
\emph{Audit.} One arm pushing the ball up a wall: 26 of 28 set-ups have no IK solution, and the other two failed in Isaac as the ball slid along the wall.
That one hand cannot take the ball rests on the grasp-width rule, which refuses a body wider than the open fingers unless another hand presses its far side.

\paragraph{Con-04, pressure test.}
\emph{``Hold the steel pipe at its test pressure for a second: both bleed valves shut while the foot valve lets air in.''}
Three Frankas; valve spring 2.0 / 3.0\,N\,m/rad.
A steel test pipe lies east--west on two stands with a pressure gauge on top, and takes its air through a hose from a foot valve in front of \Ss.
It holds its test pressure only while both bleed valves are shut and the foot valve lets air in.
Each bleed valve shuts while its lever on the pipe's top is pressed down, and a spring lifts the lever again when it is let go; the foot valve's pedal likewise rises and shuts when let go.
\emph{Reference} (7.8 / 7.9\,s): \W and \E each press their lever's grip down, \Ss presses the pedal's toe bar down, and all three hold for the test.
\emph{Success.} Both levers and the pedal are held down at once for 1\,s.
\emph{Audit.} Each lever's grip is reached only by the arm at its end of the table (\Ss is 1.04\,m from either, the far end arm 1.02\,m), and with the end arms on the levers no hand is left for the pedal; the two-arm attempt fails in Isaac in all 12 scenes.

\paragraph{Rea-01, cross-zone relay.}
\emph{``Move the apple from the start area at the far west end of the table into the crate at the far east end.''}
\W PiPER, \Ss Franka, \E PiPER; fruit 1 / 3 pieces.
Fruit starts behind \W and must end in a crate behind \E, 2.1\,m away.
A partition across the middle, 50\,cm high where the end arms reach high, keeps them from sharing any place; south of it a curb runs to the table's edge, and the only way through is a low channel roofed over its 30\,cm middle, where neither end arm reaches.
\emph{Reference} (28.7 / 48.9\,s): \W $\to$ pad A $\to$ \Ss $\to$ pad B $\to$ \E $\to$ crate.
\emph{Success.} Every piece of fruit lies in the crate for 1\,s and to the end of the run.
\emph{Audit.} The end PiPERs reach 0.75\,m, beyond their nominal 0.626\,m; the partition and the roofed middle of the channel keep them from meeting.

\paragraph{Rea-02, tomato flume.}
\emph{``Roll the cherry tomato from the top of the shelf at the west end of the table into the pot at the east end, along a track laid on the stands. Do not carry it.''}
\W PiPER, \Ss Franka, \E PiPER; fit room 6 / 3\,mm.
Three wooden channels (0.80, 0.55 and 0.80\,m, each wider than the one before) are laid end to end on two stands, every lower end inside the next upper end.
Each channel starts where only one arm reaches it, the short middle one behind \Ss, and only \W reaches the stop at the shelf's edge; laid, the channels stay put, so no arm holds anything while the tomato rolls.
\emph{Reference} (44.0\,s): \E lays the lowest channel, \Ss the middle one, \W the upper one; \W lifts the stop.
\emph{Success.} The tomato reaches the pot within 5\,s of leaving the shelf, never lying on the table, and stays there for 1\,s.
\emph{Audit.} The short middle channel sits behind \Ss, out of the end arms' reach; a cradle rules out raking the tomato along without the middle arm.

\paragraph{Rea-03, domino run.}
\emph{``Set the line of dominoes off from its west end so that it falls all the way to its east end.''}
Three PiPERs; spacing 75 / 79\,mm.
A line of 25 dominoes runs 1.8 / 1.9\,m down the table, 20\,cm in front of \Ss, with three gaps: the first place, the last, and places 12--14 in the middle; each gap is within one arm's reach only, and only \W reaches the first place.
A domino stood in its gap stays up, so no arm holds anything while another works.
\emph{Reference} (49.4 / 50.2\,s): the end arms fill the end places, \Ss the middle three; \W tips the first domino east.
\emph{Success.} The line is whole when its first domino falls, and every domino on it has fallen, tilted past 15\textdegree, for 1\,s.
\emph{Audit.} All margins hold at least 3\,cm; since the line must be whole, setting its two halves off apart does not count.

\paragraph{Rea-04, wipe the spills.}
\emph{``Wipe each of the three coffee spills on the counter with the sponge, keeping it down on each spill for 1\,s, then drop the sponge in the sink at the east end.''}
Three PiPERs; wipe time 1 / 2\,s.
Three spills lie along the counter, one in each arm's part of it, and one sponge waits on a dish at the far west end; the dish is \W's alone and the sink \E's alone.
The sponge goes from arm to arm on pads that both neighbors reach, and put down, it stays.
The middle spill lies on a 6\,cm riser 19\,cm in front of \Ss and 0.87\,m from the end arms.
\emph{Reference} (55.0 / 58.0\,s): \W wipes its spill and sets the sponge on pad A; \Ss wipes the middle spill and sets it on pad B; \E wipes its spill and drops the sponge in the sink.
\emph{Success.} The sponge stays down on each spill, its middle within 2.5\,cm of the spill's, for the wipe time, then lies in the sink for 1\,s and to the end of the run.
\emph{Audit.} The middle spill lies in front of \Ss, out of the end arms' reach (margins hold at 8.5\,cm, a hand holding the sponge by its end).

\subsection{Versions, evidence and props}
\label{sec:supp-versions}

\begin{figure*}[t]
  \centering
  \newcommand{\tm}[1]{\textsf{#1}}
  \newcommand{\vt}[2]{%
    \begin{minipage}[t]{0.193\linewidth}\centering
      \includegraphics[width=\linewidth]{figures/supp/scenes/#1.jpg}\\[-1.5pt]
      {\scriptsize\strut #2}
    \end{minipage}}
  \newcommand{\vh}[1]{\begin{minipage}[b]{0.193\linewidth}\centering\scriptsize\ver{#1}\end{minipage}}
  \newcommand{\vheads}{\vh{exact}\hfill\vh{surplus}\hfill\vh{shortfall}\hfill\vh{swapped}\hfill\vh{borrow}}
  \newcommand{\vtitle}[2]{{\scriptsize\strut\textbf{#1} #2}\\[-0.5pt]}
  \begin{minipage}[t]{0.493\textwidth}\vheads\end{minipage}\hfill
  \begin{minipage}[t]{0.493\textwidth}\vheads\end{minipage}\\[2pt]
  \begin{minipage}[t]{0.493\textwidth}\vtitle{Con-01}{spring doors}%
    \vt{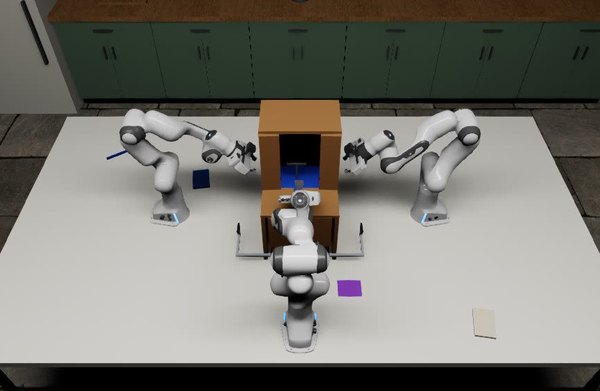}{\tm{W\,S\,E}}\hfill\vt{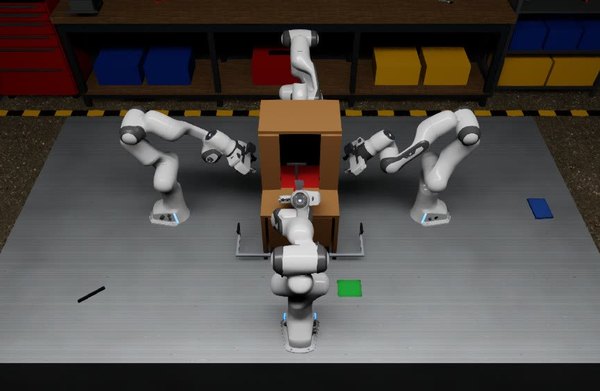}{\tm{W\,S\,E}}\hfill\vt{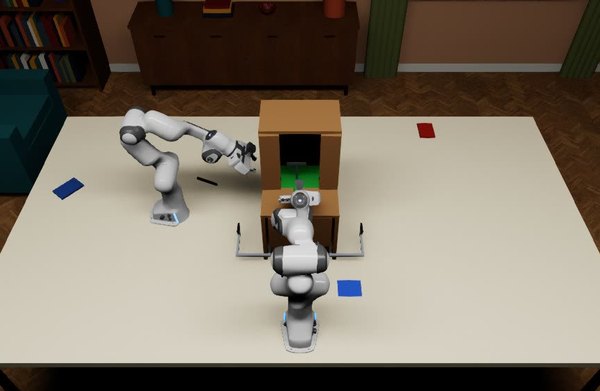}{\nope}\hfill\vt{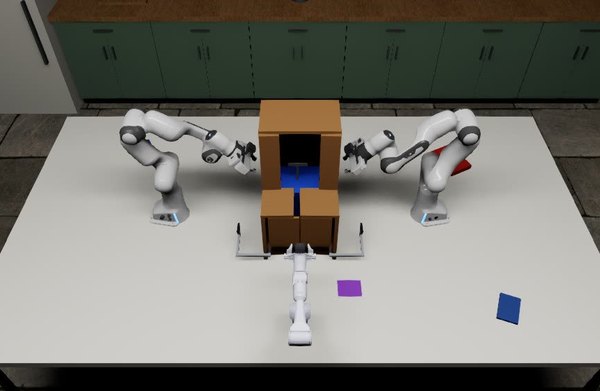}{\nope}\hfill\vt{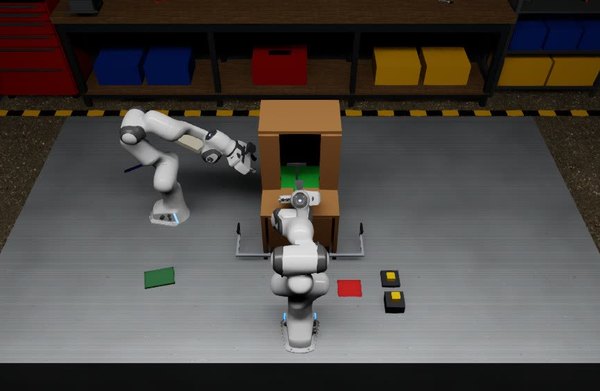}{\tm{W\,S} + weight}
  \end{minipage}\hfill
  \begin{minipage}[t]{0.493\textwidth}\vtitle{Con-02}{hinged leaves}%
    \vt{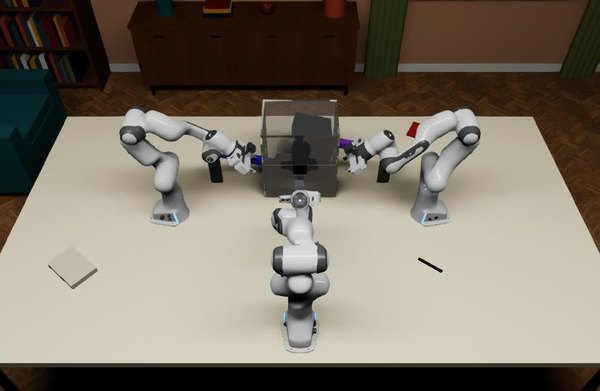}{\tm{W\,S\,E}}\hfill\vt{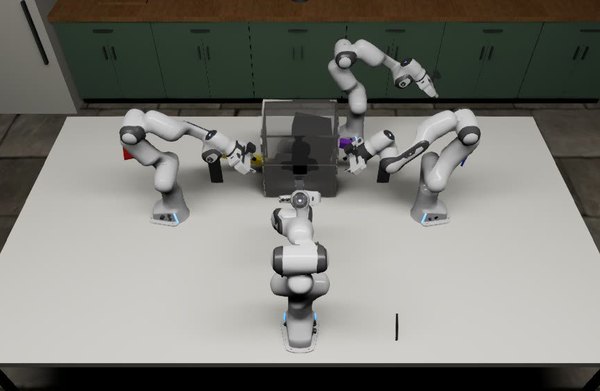}{\tm{W} + any two}\hfill\vt{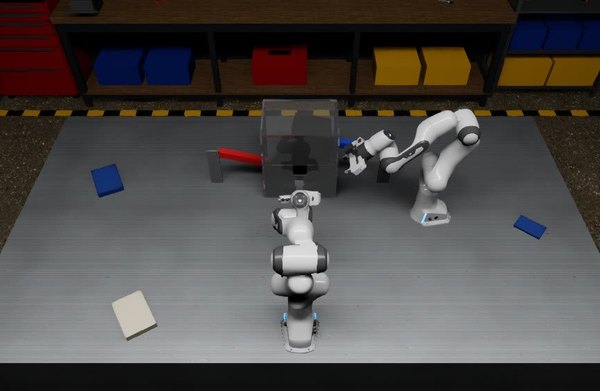}{\nope}\hfill\vt{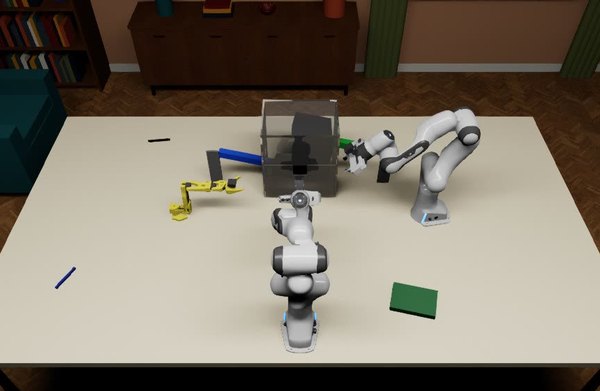}{\nope}\hfill\vt{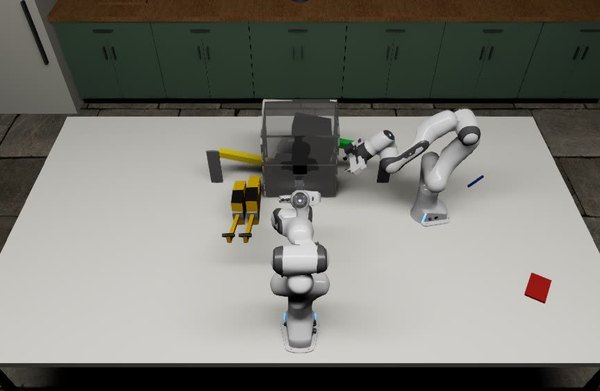}{\tm{S\,E} + chock}
  \end{minipage}\\[7pt]
  \begin{minipage}[t]{0.493\textwidth}\vtitle{Con-03}{spring lid}%
    \vt{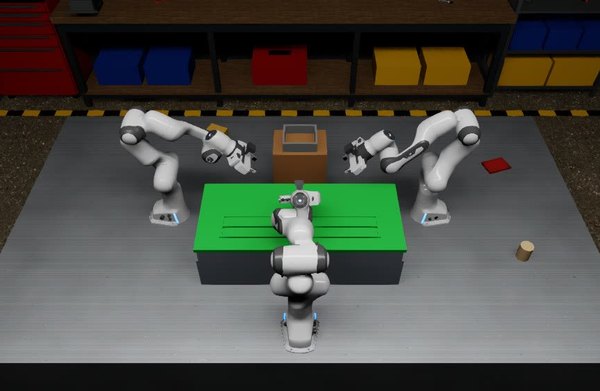}{\tm{W\,S\,E}}\hfill\vt{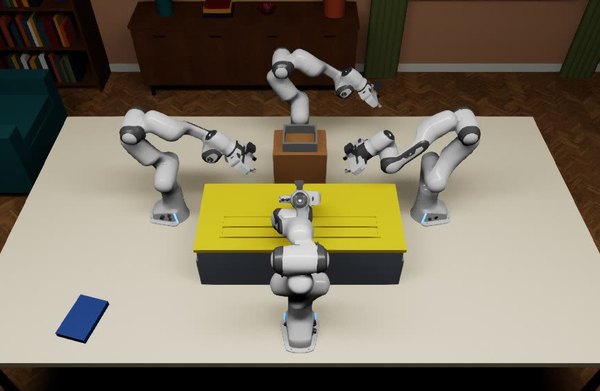}{\tm{W\,S\,E}}\hfill\vt{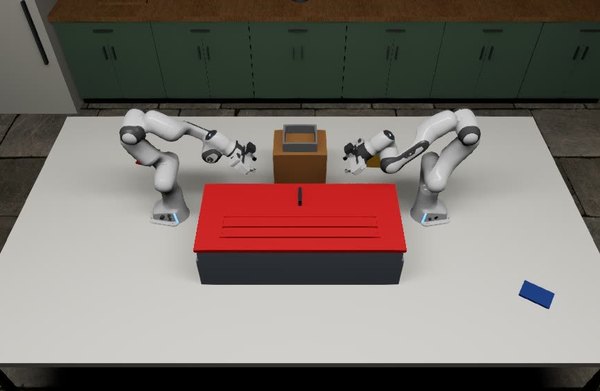}{\nope}\hfill\vt{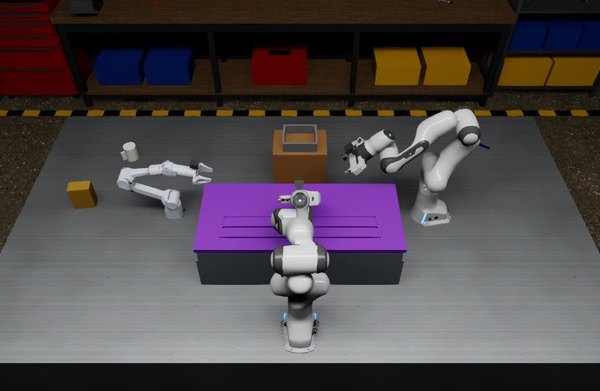}{\nope}\hfill\vt{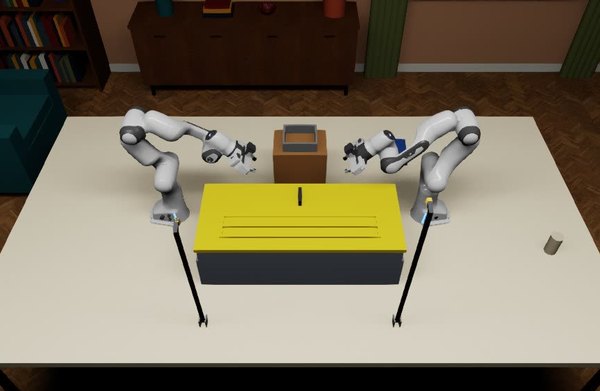}{\tm{W\,E} + stay}
  \end{minipage}\hfill
  \begin{minipage}[t]{0.493\textwidth}\vtitle{Con-04}{pressure test}%
    \vt{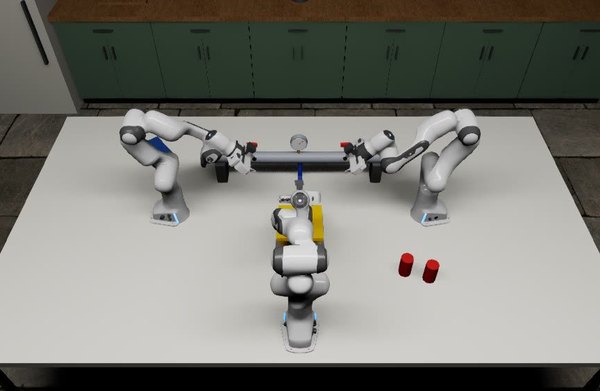}{\tm{W\,S\,E}}\hfill\vt{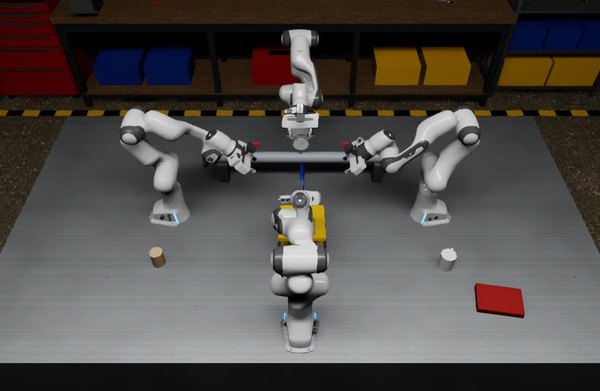}{any three}\hfill\vt{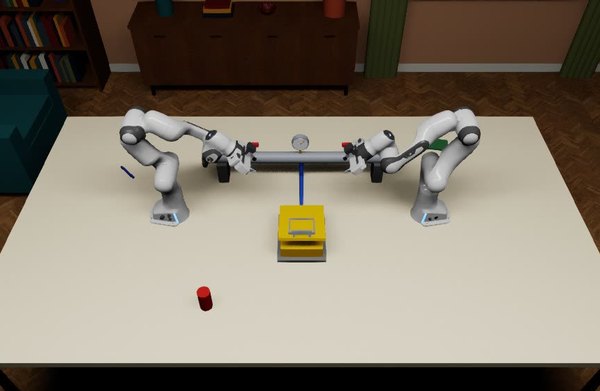}{\nope}\hfill\vt{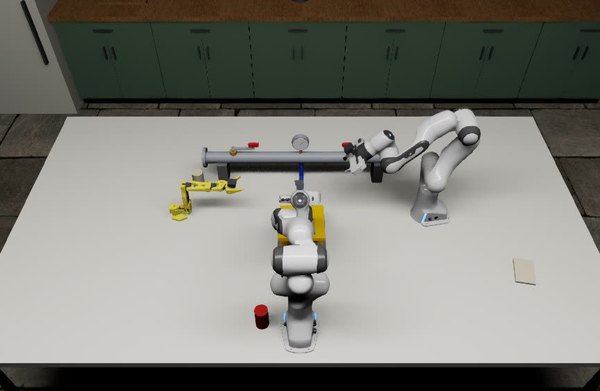}{\nope}\hfill\vt{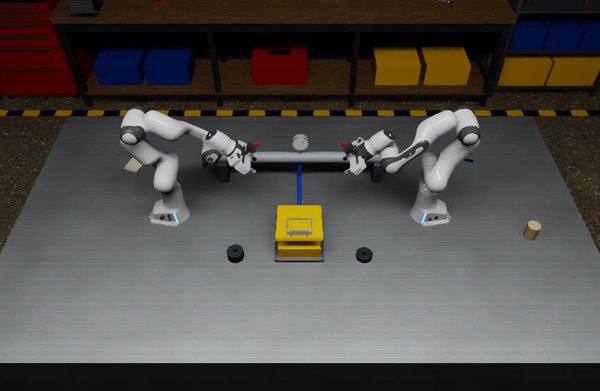}{\tm{W\,E} + weight}
  \end{minipage}\\[7pt]
  \begin{minipage}[t]{0.493\textwidth}\vtitle{Rea-01}{cross-zone relay}%
    \vt{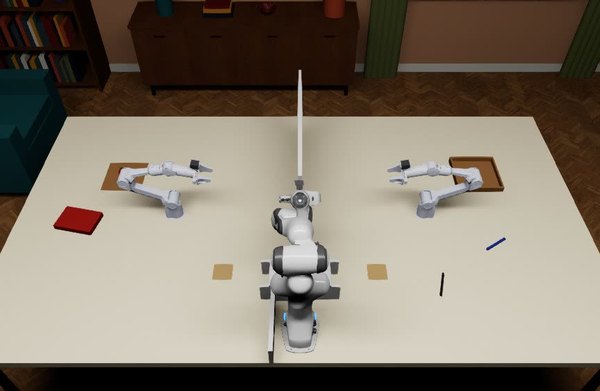}{\tm{W\,S\,E}}\hfill\vt{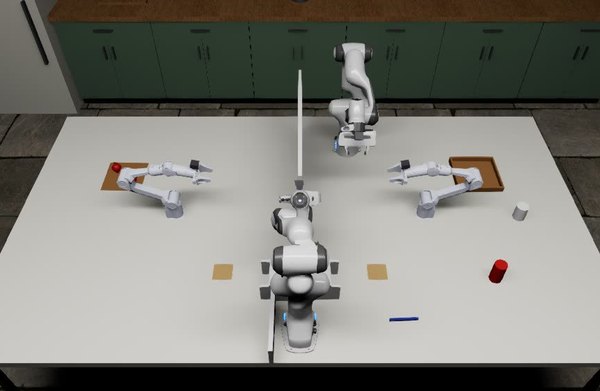}{\tm{W\,N}}\hfill\vt{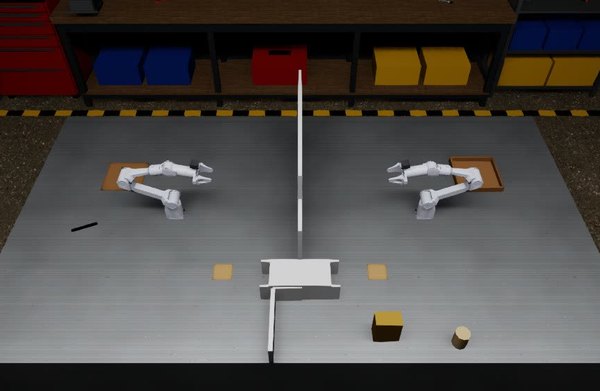}{\nope}\hfill\vt{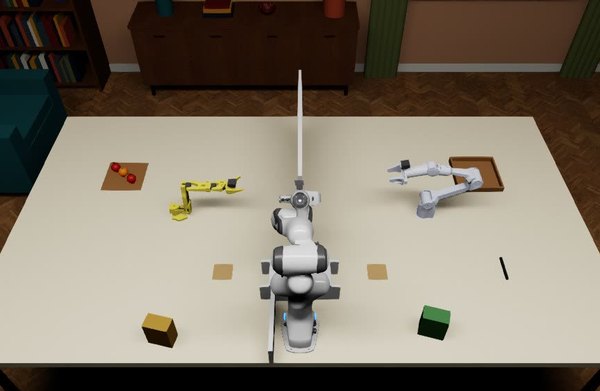}{\nope}\hfill\vt{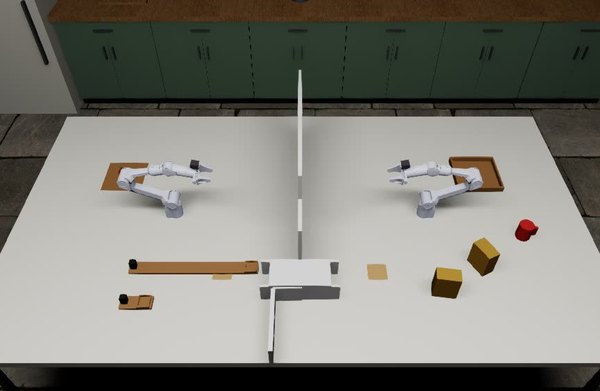}{\tm{W\,E} + paddle}
  \end{minipage}\hfill
  \begin{minipage}[t]{0.493\textwidth}\vtitle{Rea-02}{tomato flume}%
    \vt{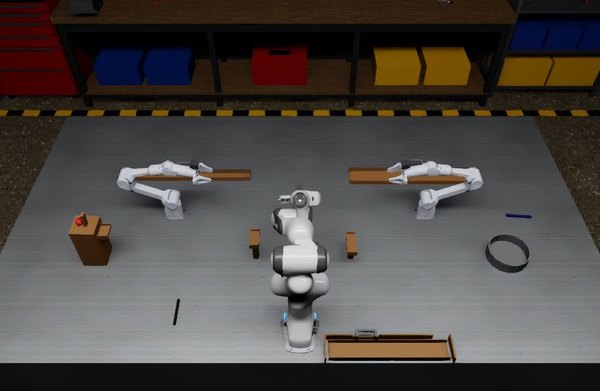}{\tm{W\,S\,E}}\hfill\vt{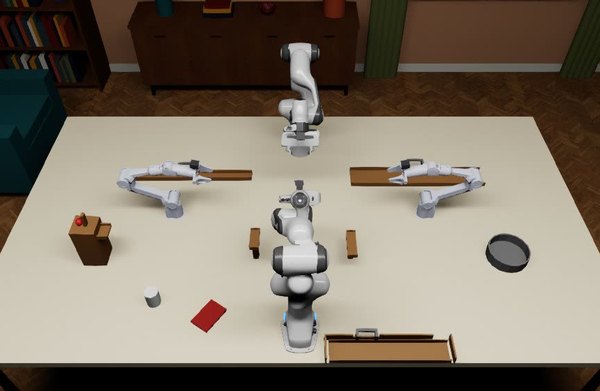}{\tm{W\,S} + \tm{E} or \tm{N}}\hfill\vt{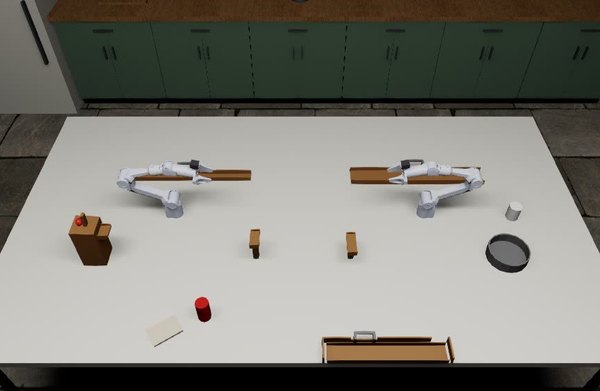}{\nope}\hfill\vt{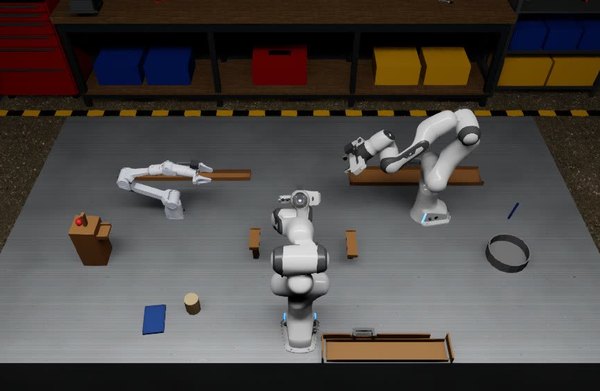}{\tm{W\,E}}\hfill\vt{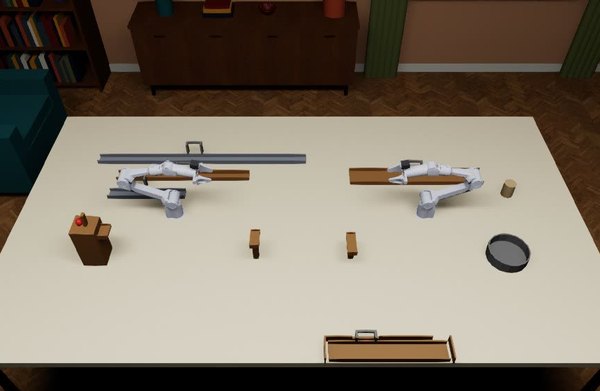}{\tm{W\,E} + chute}
  \end{minipage}\\[7pt]
  \begin{minipage}[t]{0.493\textwidth}\vtitle{Rea-03}{domino run}%
    \vt{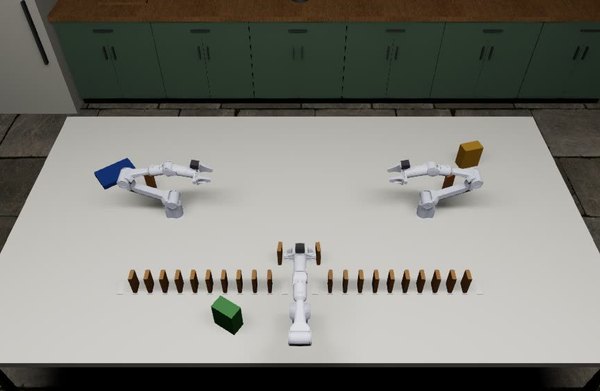}{\tm{W\,S\,E}}\hfill\vt{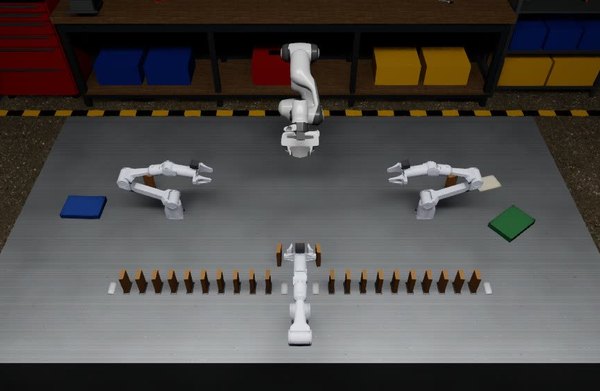}{\tm{W\,S\,E}}\hfill\vt{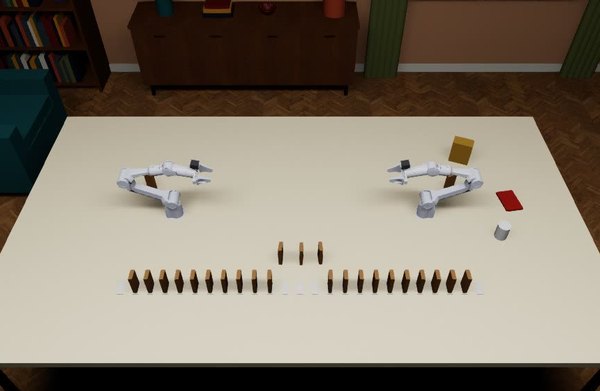}{\nope}\hfill\vt{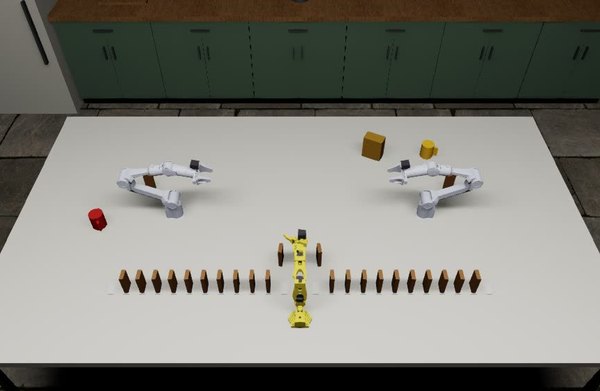}{\nope}\hfill\vt{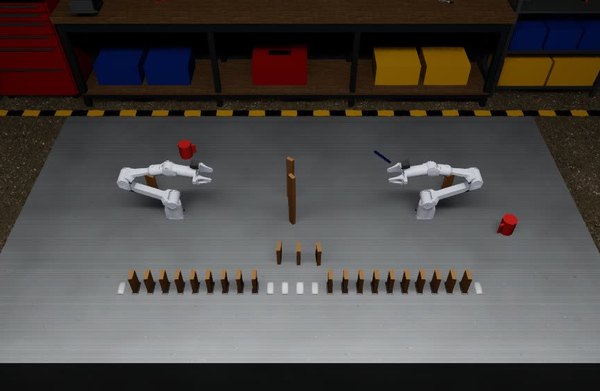}{\tm{W\,E} + slat}
  \end{minipage}\hfill
  \begin{minipage}[t]{0.493\textwidth}\vtitle{Rea-04}{wipe the spills}%
    \vt{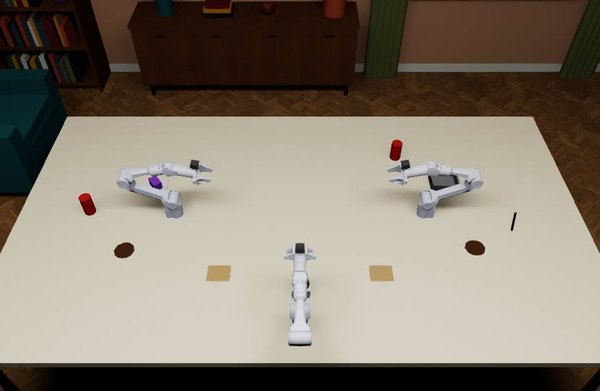}{\tm{W\,S\,E}}\hfill\vt{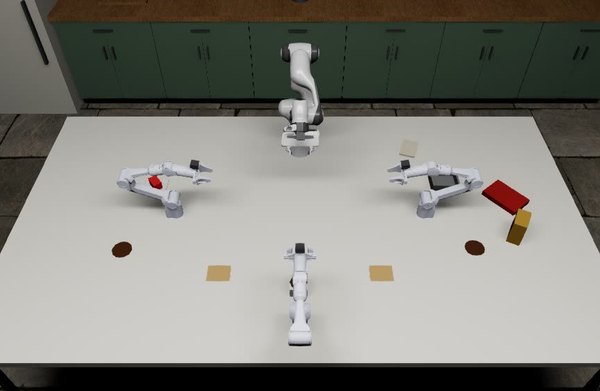}{\tm{W\,S\,E}}\hfill\vt{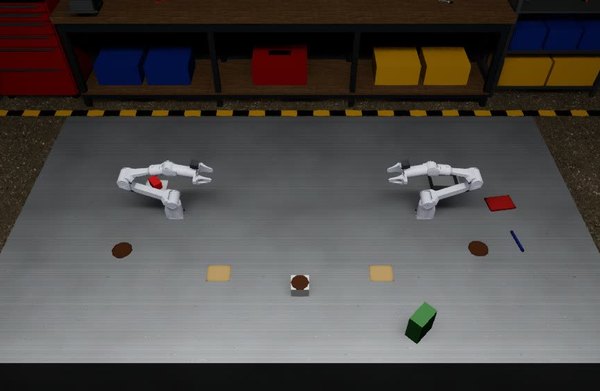}{\nope}\hfill\vt{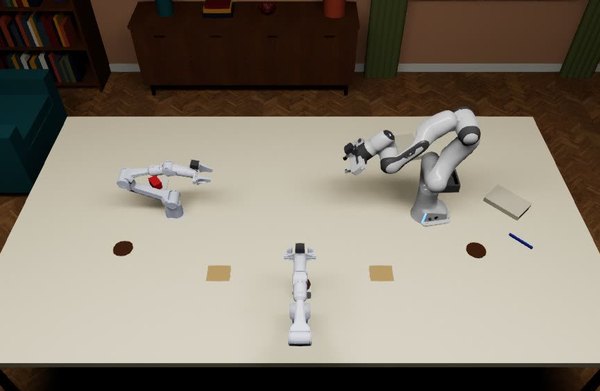}{\tm{W\,E}}\hfill\vt{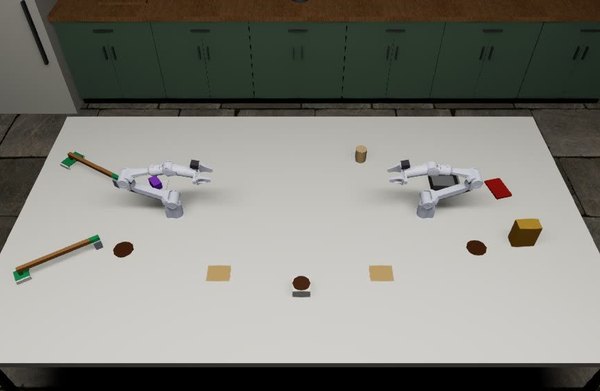}{\tm{W\,E} + brush}
  \end{minipage}
  \caption{\textbf{Every task in its five configurations}: the start of one scene each, with its gold answer below, the arms of a minimal support set (with the prop if needed) or \nope, infeasible (\cref{tab:supp-tasks}). Seed and level vary, and with them the room and the objects; the goal does not.}
  \label{fig:supp-versions}
\end{figure*}

\begin{table*}[t]
  \caption{\textbf{Tasks and gold answers.} The arms at \W/\Ss/\E (F: Franka, Pi: PiPER) and the gold answer of each version: the arms of a minimal support set (a bar separates alternatives), or \textcolor{slRed}{\nope}, infeasible, with the job left without a doer. \ver{surplus} adds a Franka at \N; \ver{shortfall} and \ver{borrow} remove the same arm, \E in Con-01, \W in Con-02 and \Ss in the other six; \ver{swapped} names the arm replaced; \ver{borrow} names the prop and, in brackets, the distractor. Knobs of the two levels are in the task cards (\cref{sec:supp-tasks}).}
  \label{tab:supp-tasks}
  \centering
  \scriptsize
  \setlength{\tabcolsep}{2.6pt}
  \renewcommand{\arraystretch}{1.12}
  \newcommand{\team}[1]{\textsf{#1}}
  \newcommand{\infe}[1]{\textcolor{slRed}{\nope}\,{\color{slGray}#1}}
  \newcommand{\famrow}[3]{\multicolumn{7}{@{}l}{\cellcolor{#1Tint}\textcolor{#1Ink}{\textbf{#2}}\ \ \textcolor{slGray}{#3}}}
  \begin{tabularx}{\textwidth}{@{}l c l >{\hsize=0.939\hsize\raggedright\arraybackslash}X >{\hsize=0.788\hsize\raggedright\arraybackslash}X l >{\hsize=1.273\hsize\raggedright\arraybackslash}X@{}}
    \toprule
    & & \multicolumn{5}{c}{Gold answer per version} \\
    \cmidrule(l){3-7}
    Task & \W/\Ss/\E & \ver{exact} & \ver{surplus} & \ver{shortfall} & \ver{swapped} & \ver{borrow} \\
    \midrule
    \famrow{famCon}{\Con}{three hands at the same moment} \\
    Con-01 spring doors   & F/F/F  & \team{W\,S\,E} & \team{W\,S\,E} & \infe{east door or box} & \infe{(\Ss{}$\to$Pi)}     & \team{W\,S} + 9\,cm weight (4\,cm) \\
    Con-02 hinged leaves  & F/F/F  & \team{W\,S\,E} & \team{W\,S\,E} $|$ \team{W\,N\,E} $|$ \team{W\,S\,N} & \infe{west leaf} & \infe{(\W{}$\to$SO-101)} & \team{S\,E} + wheel chock (low or wide) \\
    Con-03 spring lid     & F/F/F  & \team{W\,S\,E} & \team{W\,S\,E} & \infe{lid or ball}      & \infe{(\W{}$\to$Pi)}     & \team{W\,E} + 0.72\,m stay ($\pm$12\,cm) \\
    Con-04 pressure test  & F/F/F  & \team{W\,S\,E} & \team{W\,S\,E} $|$ \team{N\,S\,E} $|$ \team{W\,N\,E} $|$ \team{W\,S\,N} & \infe{three holds, two arms} & \infe{(\W{}$\to$SO-101)} & \team{W\,E} + 4.8\,cm weight (2.4\,cm) \\
    \midrule
    \famrow{famRea}{\Rea}{no fixed arm reaches every place the task goes} \\
    Rea-01 cross-zone relay & Pi/F/Pi & \team{W\,S\,E} & \team{W\,N} & \infe{carry past partition} & \infe{(\W{}$\to$SO-101)} & \team{W\,E} + 67\,cm paddle (16\,cm / tall) \\
    Rea-02 tomato flume     & Pi/F/Pi & \team{W\,S\,E} & \team{W\,S\,E} $|$ \team{W\,S\,N} & \infe{middle channel} & \team{W\,E} (\E{}$\to$F)    & \team{W\,E} + 1.30\,m chute (0.45\,m) \\
    Rea-03 domino run       & Pi/Pi/Pi & \team{W\,S\,E} & \team{W\,S\,E} & \infe{places 12--14}   & \infe{(\Ss{}$\to$SO-101)} & \team{W\,E} + 30\,cm slat (24 / 36\,cm) \\
    Rea-04 wipe the spills  & Pi/Pi/Pi & \team{W\,S\,E} & \team{W\,S\,E} & \infe{middle spill}    & \team{W\,E} (\E{}$\to$F)    & \team{W\,E} + sponge brush (clip narrow / wide) \\
    \midrule
    \rowcolor{tabGroup}\multicolumn{2}{@{}l}{Shortcut the configuration defeats} & \textcolor{slGray}{--} & \textit{staffing by presence} & \textit{presumed feasibility} & \textit{roles by place} & \textit{substitution by appearance} \\
    \bottomrule
  \end{tabularx}
\end{table*}

\input{figures/supp/supp-borrow}

\begin{table*}[t]
  \caption{\textbf{Props and distractors.} Each \ver{borrow} version removes one arm; its prop does that arm's job, and its distractor is the same kind of object that differs in one measurable property (Con-04's, half as tall, is also half as heavy). Masses are given only roughly, so an object must be measured against its job.}
  \label{tab:supp-props}
  \centering
  \footnotesize
  \setlength{\tabcolsep}{4pt}
  \renewcommand{\arraystretch}{1.15}
  \newcommand{\famrow}[3]{\multicolumn{5}{@{}l}{\cellcolor{#1Tint}\textcolor{#1Ink}{\textbf{#2}}\ \ \textcolor{slGray}{#3}}}
  \newcommand{\tk}[2]{\makecell[tl]{#1\\[-1pt]\textcolor{slGray}{#2}}}
  \begin{tabularx}{\textwidth}{@{}l >{\raggedright\arraybackslash}p{2.45cm} >{\raggedright\arraybackslash}X >{\raggedright\arraybackslash}X l@{}}
    \toprule
    Task & Removed arm's job & Prop & Distractor & Differs in \\
    \midrule
    \famrow{famCon}{\Con}{the prop holds what the removed arm held} \\
    \tk{Con-01}{spring doors} & \E: hold the east door open & a 9\,cm counterweight in the east lever's way back: the bar, 5\,cm up, cannot pass it & 4\,cm tall: the lever passes over it and the door shuts & height \\
    \tk{Con-02}{hinged leaves} & \W: hold the blue leaf level & a wheel chock pushed under the hanging bar; its top sits 0.8\,mm below the level bar & 1.9\,cm lower (the bar rests 4\textdegree{} low or more), or 32\,cm wide (the gap is 25.5\,cm) & height, width \\
    \tk{Con-03}{spring lid} & \Ss: hold the lid open & a rod on a pivoting foot, its hook 0.72\,m out, swung over the lid's edge & 12\,cm longer (the hook passes the free edge) or 12\,cm shorter (it never reaches it) & length \\
    \tk{Con-04}{pressure test} & \Ss: hold the foot valve down & a test weight, 4.8\,cm tall and 2.4\,kg, in the pedal's pocket: it holds the pedal down & half as tall and as heavy, 2.4\,cm and 1.2\,kg: the pedal does not move & height, mass \\
    \midrule
    \famrow{famRea}{\Rea}{the prop covers the middle the removed arm reached} \\
    \tk{Rea-01}{cross-zone relay} & \Ss: carry fruit across the middle & a 67\,cm paddle, pushed by \W through the channel until the fruit is within \E's reach & 16\,cm (the fruit stays out of both reaches), or a 12\,cm rim (the channel is 11\,cm) & length, height \\
    \tk{Rea-02}{tomato flume} & \Ss: lay the middle channel & a 1.30\,m chute, laid by \W from the shelf past the first stand into the lowest channel & 0.45\,m: it ends far short of the lowest channel & length \\
    \tk{Rea-03}{domino run} & \Ss: fill places 12--14 & a 30\,cm slat stood in place 11; it falls across places 12--14 onto the domino in place 15 & 24\,cm (it falls into the gap) or 36\,cm (it presses the domino in place 15 down) & length \\
    \tk{Rea-04}{wipe the spills} & \Ss: wipe the middle spill & a sponge brush whose clip holds the sponge; \W lifts it onto the riser and wipes & a clip too narrow (refused by the grasp rule) or too wide (it only pushes the sponge) & clip width \\
    \bottomrule
  \end{tabularx}
\end{table*}

\begin{table}[t]
  \caption{\textbf{State of the 40 versions.} I: the gold answer succeeds in Isaac in all 12 scenes of the version (both levels, seeds 0--5; each set where several are right), and in \ver{borrow} the distractor fails in all 12; S: infeasibility shown by the solver within the job model (IK with margins).}
  \label{tab:supp-state}
  \centering
  \small
  \setlength{\tabcolsep}{2pt}
  \begin{NiceTabular*}{\linewidth}{@{\extracolsep{\fill}}lccccc@{}}[cell-space-limits=1pt]
    \toprule
    Task & \ver{exact} & \ver{surplus} & \ver{shortfall} & \ver{swapped} & \ver{borrow} \\
    \midrule
    Con-01 & I & I & S & S & I \\
    Con-02 & I & I (each set) & S & S & I \\
    Con-03 & I & I & S & S & I \\
    Con-04 & I & I (each set) & S & S & I \\
    Rea-01 & I & I & S & S & I \\
    Rea-02 & I & I (each set) & S & I & I \\
    Rea-03 & I & I & S & S & I \\
    Rea-04 & I & I & S & I & I \\
    \bottomrule
  \end{NiceTabular*}
\end{table}

\paragraph{Answers per version.}
\Cref{tab:supp-tasks} gives each version's gold answer and \cref{tab:supp-state} its evidence, and \cref{fig:supp-versions} shows every version of every task at the start of a varied scene.
In \ver{surplus} a Franka stands at \N.
Its answer is the three arms of \ver{exact} with \N idle in Con-01, Con-03, Rea-03 and Rea-04; one of several alternative sets in Con-02, Con-04 and Rea-02 (in Rea-02, \N carries the lowest channel to \Ss, and \E stays idle); and \W{}+\N in Rea-01, where \W hands each piece to \N west of the tall partition and \Ss and \E stay idle.
In \ver{swapped} the replacement is a PiPER or an SO-101 where a Franka stood, an SO-101 where a PiPER stood, or a Franka where a PiPER stood; six swaps make the task infeasible, and in Rea-02 and Rea-04 an east Franka reaches the middle arm's work, so that \W{}+\E suffice.
In Rea-04 the east Franka takes the sponge by its end, the only grip that reaches the riser 0.87\,m out.

\paragraph{What the gold answers rest on.}
Every gold answer is checked (\cref{tab:supp-state}): the gold teams succeed in all 312 feasible scenes, and the distractors fail in all 96 \ver{borrow} scenes.
A reference that succeeds in Isaac shows that its team suffices; the solver, which tries teams from the smallest up under IK with margins, shows that no smaller team suffices and, in the infeasible versions, that the jobs cannot all have a doer.
In the main runs, each of the 306 answers that name fewer arms than the gold answer on a feasible scene was carried out, and none succeeded.
In each two-arm attempt, the two arms of the \ver{shortfall} team replay the scripted reference with the missing arm's actions removed, after one takes over that arm's hold; in all 96 task-scenes, every attempt fails, and shows the missing job's signal.
Each task card's audit (\cref{sec:supp-tasks}) gives why no two arms suffice.
In Con-04 \ver{shortfall} each hold has an arm, but the three must happen at once and two arms remain.

\paragraph{Props and distractors.}
\Cref{tab:supp-props} lists each pair.
A prop and its distractor are the same kind of object and differ in one measurable geometric property, never in mass alone or in a hidden state; Con-04's distractor, half as tall, is also half as heavy. Where a task has two distractors, each differs from the prop in one property.
Wherever a larger distractor that fails exists, the seed sets whether the prop or the distractor is the larger; in the remaining scenes the prop is the larger, which the fixed rules of \cref{tab:supp-rules} exploit.
\Cref{fig:supp-borrow} plays each \ver{borrow} reference with its prop and with its distractor: the prop does the removed arm's job, and the distractor fails by that job's signal.

\paragraph{What the seeds change.}
Within a version and level, the seed sets the room, the colors and three everyday objects, shifts the task's objects by at most 1.6\,cm, and in \ver{borrow} draws where the prop and the distractor stand and which is larger.
The 480 scenes therefore hold 80 distinct physical problems (40 versions $\times$ 2 levels), each in six seeds that change the appearance and shift the objects by at most 1.6\,cm: the seeds test robustness to appearance and placement.

\subsection{Tools, program API and isolation}
\label{sec:supp-tools}

\begin{table}[t]
  \caption{\textbf{Shared tools.} Every method reaches them through one environment server; an arm moves only through the program API. Per episode, \texttt{try\_in\_sim} runs at most three times, in the main runs as under the equal budget, where \texttt{probe} runs at most twice.}
  \label{tab:supp-tools}
  \centering
  \scriptsize
  \setlength{\tabcolsep}{3pt}
  \renewcommand{\arraystretch}{1.05}
  \begin{tabularx}{\linewidth}{@{}l >{\raggedright\arraybackslash}X@{}}
    \toprule
    Tool & What it returns \\
    \midrule
    \multicolumn{2}{@{}l}{\textbf{Senses and measurements}} \\
    \texttt{briefing}    & goal; arm models and poses; rough masses; fixed objects \\
    \texttt{look}        & one image from a fixed or a wrist camera \\
    \texttt{read\_arm}   & joint angles and limits, tool pose, finger opening, torques \\
    \texttt{locate}      & an object's or a part's box: the simulator's, with noise of 4\,mm and 1.5\textdegree \\
    \texttt{measure}     & an object's parts and sizes: lengths, heights, openings \\
    \texttt{heft}        & mass to a few per cent, by an arm in reach (computed) \\
    \texttt{can\_reach}  & own IK at a pose: margin or distance short (1--2\,cm) \\
    \texttt{who\_can}    & every arm's margin at a point \\
    \texttt{meet}        & where two arms can hand over, above an obstacle too \\
    \texttt{scene\_graph}& the objects in the scene and their relations \\
    \texttt{check\_hold} & whether a hand, a stop or a weight can hold an object still \\
    \texttt{note}        & a notebook of confirmed facts and their evidence \\
    \texttt{probe}       & a motion of at most 8\,s, played in Isaac and read back \\
    \multicolumn{2}{@{}l}{\textbf{Control}} \\
    program API          & \texttt{tool\_pose}, \texttt{above}, \texttt{shift}, \texttt{back}, \texttt{where}, \texttt{move}, \texttt{grip}, \texttt{push}, \texttt{release}, \texttt{open\_fingers}, \texttt{phase}; a program sets one \texttt{PLAN} and imports only \texttt{math} \\
    \texttt{check\_program} & conflicts in a sandbox run: IK on all paths; clearance 3\,cm between arms, 1\,cm to the scene, 1.5\,cm for a held object \\
    \texttt{try\_in\_sim} & a physics run (joint targets, privileged attachment within 2\,cm after the width check, cloth pinching): four keyframes, event log, success, failure report \\
    \multicolumn{2}{@{}l}{\textbf{Output}} \\
    \texttt{submit\_team\_plan} & feasible or not; arms used, idle arms and why; objects used and their jobs, rejected with evidence; what is missing \\
    \texttt{submit\_program} & the final program \\
    \texttt{respond} & a session's structured reply within its row's procedure: subtasks, coalitions, a turn of dialogue, a check \\
    \multicolumn{2}{@{}l}{\textbf{LLM + solver only}} \\
    \texttt{solve\_team} & exact search over a job graph: every smallest team and who does what, or what blocks it \\
    \bottomrule
  \end{tabularx}
\end{table}

\Cref{tab:supp-tools} lists the shared tools.
Three boundaries keep them fair.
\texttt{heft} is a measurement like \texttt{measure}: it does not count against the probes, and it must name an arm that reaches the object.
Each session gets the tools its method's design gives it (\cref{sec:supp-methods}).
Rough masses are an observation that every row receives: the true mass times a factor in 0.75--1.3, the same in every version of a task.
In rows that decide per arm, every arm sees the fixed cameras, but only its own wrist camera, \texttt{read\_arm} and \texttt{can\_reach}.

\paragraph{What the tools know.}
No tool reads the jobs, the roles, the success predicates or the scripted reference; of the card, \texttt{briefing} gives the goal and what does not show, such as a spring, and a painted place such as Con-01's mat is a scene object that \texttt{locate} returns. Every point a method asks about is its own, taken from the boxes \texttt{locate} returns.
\texttt{can\_reach} and \texttt{who\_can} answer from the arms' kinematics, checked against the table and fixed bodies but not along a path, against other arms or under a payload; \texttt{meet} searches between two arms' bases for places where both reach an object from their own sides, clear of fixed bodies; \texttt{check\_hold} computes by statics, with the rough masses, whether the holds a method lists keep a body still against its weight and any spring.
None says what an object is for or which job it serves.
The failure report, which names the jobs whose signals fired, comes only from \texttt{try\_in\_sim}, and in the one-shot setting the team is submitted before any run.

\paragraph{Isolation.}
Every session is started without access to the benchmark's code: either through a command-line agent with all of its own tools switched off, or, for the main table, through the model's API with our own loop.
In both, the only tools are the environment's, and the context holds only the method's prompts, the tools and their replies.

\paragraph{An episode, step by step.}
In Rea-01 the agent receives only the goal and the briefing.
It looks at the table and locates the apple, the crate, the partition and the pads; it writes the jobs (take the apple at the start, carry it past the partition, put it in the crate); it asks who reaches each site, finding that only \W reaches the start, only \E the crate, and that the partition leaves no place both end arms reach; and it submits \W{}+\Ss{}+\E with a reason for each arm.
The shared executor then writes a program in phases for the declared arms, \texttt{check\_program} checks its paths, and \texttt{try\_in\_sim} runs it; after a failed run the executor may write a new program, up to three runs, but nothing goes back to the method.
The scored team is the declared arms plus any arm that moved.

\subsection{Baselines in detail}
\label{sec:supp-methods}

\paragraph{Scope of the main paper's table of closest work.}
The table scores reported formulations and evaluations, not what a method could do.
Heterogeneous arms are different manipulator models.
Selecting a team includes dynamic activation and need not give the fewest robots; a robot it leaves out sits out the whole task, not a temporary \texttt{WAIT}.
An infeasible answer is a verdict on the whole task, not a failed rollout.
A passive substitute takes over an arm's support role.
Matched answer changes keep the instruction while changing the gold team or the feasibility answer; ours are defined by numeric margins (slack against a physical bound) rather than by object dimensions, error messages or a Boolean IK result, and none of the listed works reports changes of this kind.
Physics execution uses dynamics or hardware.
For Tool-RoCo, the arm-model and execution cells also draw on its released code.
Its robots stay inactive until an agent activates them, so one can sit out a whole run; but the four reported settings in which none was activated all fail, and a robot left out is not scored.
SMART-LLM's released allocation prompt asks for the fewest robots needed, but the paper does not score minimum teams.
EMOS executes plans geometrically, with physics and collision detection turned off, and calls its example of an unassigned robot inefficient instead of showing that the smaller team succeeds.
The heterogeneous-arm cell of VIKI-Bench, the benchmark of VIKI-R, rests on a hardware case study that selects among arm models.

\paragraph{Why adapt.}
SMART-LLM can leave robots out and refuses in prose tasks whose skills are missing, but it does not score infeasibility.
Its released prompt already asks for the fewest robots.
RoCo's \texttt{WAIT} pauses an arm only for a step, and its broken-gripper example concerns the rest of the task.
EMOS tests reach quantitatively and can leave robots unassigned, but reports no feasibility slack.
HMAS-2 checks each assigned action but has no objective over the whole team.
VIKI-R scores selected sets and plans, not physical task success, and its box-support example does not show that an arm can be left out.
Two UR5e arms share a task in RoCo, HMAS-2 runs three or six arms of one kind, and one VIKI-R example activates two Panda arms; EMOS teams robots of four different types.
RoCo, EMOS and VIKI-R revise a plan only from checks made before anything is executed: validation feedback, each robot's check against its resume, and a plan checker.
HMAS-2 also re-plans when an executed action fails, as when a box slips from a gripper.
Our adaptations give them the common outputs and passive candidates of \Bench.

\paragraph{Revision after a failed run.}
Under the equal budget, every row receives the same failure report and may change its team.
SMART-LLM-style puts the last plan and the report into the prompt of its first stage, decomposition, and reruns all three stages.
RoCo-style sends the report to all arms as an environment message and opens a new round of dialogue; as in the first, the latest proposal is submitted once all agree or after three rounds.
HMAS-2-style sends the report to the central model and to every arm and opens a new round of proposal and veto; a plan still vetoed after three rounds is submitted as it stands.
LLM + solver's search takes arms, passive objects and transfers between arms. After a failed run a new session receives the last plan and the report, writes the job graph again, and the solver solves again.
In every row, a failure of execution alone may be repaired by changing only the program.

\paragraph{Fixed rules.}
The rules of \cref{tab:supp-rules} are scored against the gold answers of all 480 scenes and use only the arms present, their models and the objects' sizes: ``arm count, larger object'' answers \W{}+\Ss{}+\E whenever three or more arms are present, the two arms with the larger of the two look-alike objects when two arms and such a pair are present, and infeasible otherwise; each indented row adds a clause that answers infeasible, when an SO-101 is present and then when two Frankas stand with a PiPER.

\paragraph{Rules of comparison.}
Every row uses the same backbone within a table (Qwen3.8-Max or Claude Sonnet 5.5, at high reasoning effort), the same information, the same tools divided among each row's sessions by its design, the same output form, the same probes and runs, and the same tuning effort.
Grasps are by attachment in every row, after the same width check.

\subsection{Scoring}
\label{sec:supp-scoring}

\begin{table}[t]
  \caption{\textbf{Fixed rules.} Scenes right against the computed gold answers, of 96 per configuration and of 480 in All; Task: task-scenes right, of 96. An indented rule adds a clause to the rule above it.}
  \label{tab:supp-rules}
  \centering
  \footnotesize
  \setlength{\tabcolsep}{0.5pt}
  \begin{NiceTabular*}{\linewidth}{@{\extracolsep{\fill}}lccccccc@{}}[cell-space-limits=1pt]
    \toprule
    Rule & \ver{exact} & \ver{surplus} & \ver{shortfall} & \ver{swapped} & \ver{borrow} & All & Task \\
    \midrule
    Always infeasible             & 0 & 0 & 96 & 72 & 0 & 168 & 0 \\
    Always \W{}+\Ss{}+\E          & 96 & 84 & 0 & 0 & 0 & 180 & 0 \\
    Arm count, larger object      & 96 & 84 & 96 & 0 & 67 & 343 & 0 \\
    \quad + SO-101: infeasible & 96 & 84 & 96 & 48 & 67 & 391 & 25 \\
    \bottomrule
  \end{NiceTabular*}
\end{table}

\paragraph{Per configuration.}
On \ver{exact} the answer must be the gold team; on \ver{surplus}, any fewest-arm set of the four arms; on \ver{shortfall}, infeasible, with the missing job reported but not scored; on \ver{swapped}, the solver's answer (two arms in Rea-02 and Rea-04, infeasible in the other six); on \ver{borrow}, two arms and the prop, with the distractor unused.
Objects are scored only in \ver{borrow}: the answer must name the prop and must not name the distractor, and any other object it names is not scored.
The gold answer lists every smallest team (\cref{tab:supp-tasks}), and any one is right; an arm in one need not be in all, as in Con-04 \ver{surplus}, whose four smallest teams share no arm.

\paragraph{The team used.}
The team scored is the declared arms together with every arm that moves in a run that succeeds; since \texttt{check\_program} refuses a program that moves an arm the answer leaves idle, in the main runs it is the declared team.
In the one-shot setting the answer comes before any run, so the team scored is the one declared.
A team smaller than gold that succeeds is graded right, and the task's gold answer is rechecked.

\paragraph{Task level and fixed rules.}
A task-scene is right when all five of its configurations are.
\Cref{tab:supp-rules} gives the fixed rules. Per configuration they earn up to 100\%, and the rule of the main text is right in 391 of the 480 scenes (81.5\%); at the task level they earn nothing unless they read the arm models: the SO-101 is brought in only where it ends the task, which gives that rule 25 task-scenes.
\Scissors is above every rule on both backbones, over all scenes (85.4\% and 91.2\%) and at the task level (43.8\% and 59.4\%).
No rule gets right the three versions whose answer is a proper subset of the arms present, Rea-01 \ver{surplus} and Rea-02 and Rea-04 \ver{swapped}: on their 36 scenes \Scissors is right in 33 with either backbone, the other methods in 20 to 26 with Qwen3.8-Max and 22 to 29 with Claude Sonnet 5.5.

\paragraph{Errors and statistics.}
Each wrong answer is counted under one kind: an extra arm, a missing arm, another team of the fewest arms, feasibility reversed (a team given to an infeasible task, or a feasible one refused), a distractor used, or a prop missed (the prop left out where the team needs it); a failed run taken as evidence without a job's failure signal, the protocol's last class, cannot occur in one shot.
In the main runs every method answers each of the 480 scenes once per backbone, in the one-shot setting; we report means with 95\% intervals that resample the 96 task-scenes, methods compared paired by scene, and the accuracy on the held-out seeds.
Sessions count the model sessions that give the answer, the form check's included, not the executor's, which carry it out.

\subsection{Main runs in detail}
\label{sec:supp-more}

This section's tables and figures extend the paper's main results, on the same episodes unless a caption says otherwise.

\begin{table}[t]
  \caption{\textbf{\Scissors minus each method.} In points, paired by scene, over all episodes and at the task level, with 95\% bootstrap intervals that resample the 96 task-scenes, each with its five configurations (10\,000 resamples); a point is the difference of the two accuracies as the main text prints them.}
  \label{tab:supp-ci}
  \centering
  \footnotesize
  \setlength{\tabcolsep}{1.5pt}
  \begin{NiceTabular*}{\linewidth}{@{\extracolsep{\fill}}l*{4}{c}@{}}[cell-space-limits=1pt]
    \toprule
    \Block[l]{2-1}{\Scissors minus} & \Block{1-2}{Qwen3.8-Max} & & \Block{1-2}{Claude Sonnet 5.5} & \\
    \cmidrule(lr){2-3} \cmidrule(lr){4-5}
    & All & Task & All & Task \\
    \midrule
    LLM + solver             & \makecell{19.2\\[-1.5pt]{\scriptsize[13.8, 24.6]}} & \makecell{33.4\\[-1.5pt]{\scriptsize[20.8, 44.8]}} & \makecell{13.1\\[-1.5pt]{\scriptsize[9.2, 17.1]}} & \makecell{40.6\\[-1.5pt]{\scriptsize[27.1, 54.2]}} \\
    SMART-LLM-style          & \makecell{12.9\\[-1.5pt]{\scriptsize[8.3, 17.7]}} & \makecell{30.3\\[-1.5pt]{\scriptsize[16.7, 42.7]}} & \makecell{9.5\\[-1.5pt]{\scriptsize[5.4, 13.8]}} & \makecell{27.1\\[-1.5pt]{\scriptsize[13.5, 40.6]}} \\
    RoCo-style               & \makecell{24.4\\[-1.5pt]{\scriptsize[19.2, 29.6]}} & \makecell{41.7\\[-1.5pt]{\scriptsize[31.2, 52.1]}} & \makecell{15.0\\[-1.5pt]{\scriptsize[11.2, 19.0]}} & \makecell{37.5\\[-1.5pt]{\scriptsize[25.0, 50.0]}} \\
    HMAS-2-style             & \makecell{10.4\\[-1.5pt]{\scriptsize[5.8, 14.8]}} & \makecell{26.1\\[-1.5pt]{\scriptsize[14.6, 37.5]}} & \makecell{12.4\\[-1.5pt]{\scriptsize[8.3, 16.7]}} & \makecell{27.1\\[-1.5pt]{\scriptsize[13.5, 40.6]}} \\
    \bottomrule
  \end{NiceTabular*}
\end{table}

\begin{table*}[t]
  \caption{\textbf{Robustness of the main runs.} Team correctness (\%) by level, by room (kitchen, workshop, home: seeds 0 and 3, 1 and 4, 2 and 5), \ver{swapped} by the arm brought in, and \ver{borrow} by whether the prop or the distractor is the larger object, where ``take the larger'' fails.}
  \label{tab:supp-robust}
  \centering
  \small
  \setlength{\tabcolsep}{2pt}
  \begin{NiceTabular*}{\textwidth}{@{\extracolsep{\fill}}l*{9}{c}@{}}[cell-space-limits=1pt]
    \CodeBefore
      \rowcolor{tabGroup}{3,9}
    \Body
    \toprule
    \Block[l]{2-1}{Method} & \Block{1-2}{Level} & & \Block{1-3}{Room} & & & \Block{1-2}{\ver{swapped}} & & \Block{1-2}{\ver{borrow}} & \\
    \cmidrule(lr){2-3} \cmidrule(lr){4-6} \cmidrule(lr){7-8} \cmidrule(lr){9-10}
    & easy & hard & kitchen & workshop & home & SO-101 in & other & prop larger & prop smaller \\
    \midrule
    \Block[l]{1-10}{\textbf{Qwen3.8-Max}} \\
    LLM + solver             & 64.2 & 68.3 & 63.1 & 66.9 & 68.8 & 45.8 & 54.2 & 52.2 & 48.3 \\
    SMART-LLM-style          & 74.6 & 70.4 & 70.6 & 75.6 & 71.2 & 60.4 & 45.8 & 49.3 & 48.3 \\
    RoCo-style               & 62.9 & 59.2 & 63.8 & 60.6 & 58.8 & 37.5 & 56.2 & 34.3 & 37.9 \\
    HMAS-2-style             & 74.2 & 75.8 & 76.2 & 75.0 & 73.8 & 60.4 & 68.8 & 47.8 & 44.8 \\
    \midrule
    \textbf{\Scissors (ours)}   & \textbf{89.2} & \textbf{81.7} & \textbf{84.4} & \textbf{83.8} & \textbf{88.1} & \textbf{77.1} & \textbf{89.6} & \textbf{56.7} & \textbf{65.5} \\
    \midrule
    \Block[l]{1-10}{\textbf{Claude Sonnet 5.5}} \\
    LLM + solver             & 75.4 & 80.8 & 79.4 & 77.5 & 77.5 & 68.8 & 68.8 & 56.7 & 51.7 \\
    SMART-LLM-style          & 79.2 & 84.2 & 83.1 & 81.2 & 80.6 & 75.0 & 75.0 & 59.7 & 58.6 \\
    RoCo-style               & 75.4 & 77.1 & 73.8 & 77.5 & 77.5 & 77.1 & 66.7 & 44.8 & \textbf{72.4} \\
    HMAS-2-style             & 80.8 & 76.7 & 76.9 & 79.4 & 80.0 & 64.6 & 79.2 & 47.8 & \textbf{72.4} \\
    \midrule
    \textbf{\Scissors (ours)}   & \textbf{91.2} & \textbf{91.2} & \textbf{91.9} & \textbf{93.8} & \textbf{88.1} & \textbf{100.0} & \textbf{91.7} & \textbf{67.2} & 65.5 \\
    \bottomrule
  \end{NiceTabular*}
\end{table*}

\begin{table*}[t]
  \caption{\textbf{Wrong answers by kind in the main runs.} Of 480 episodes in each row, one kind per wrong answer (\cref{sec:supp-scoring}); \emph{Team given} and \emph{Refused} are the two directions of feasibility reversed, an infeasible task given a team and a feasible task refused; \emph{Other team}: another team of the fewest arms. No answer is missing in the main runs.}
  \label{tab:supp-errors-main}
  \centering
  \small
  \setlength{\tabcolsep}{2pt}
  \begin{NiceTabular*}{\textwidth}{@{\extracolsep{\fill}}lcccccccc@{}}[cell-space-limits=1pt]
    \CodeBefore
      \rowcolor{tabGroup}{2,8}
    \Body
    \toprule
    Method & Wrong & Extra arm & Missing arm & Team given & Refused & Distractor used & Prop missed & Other team \\
    \midrule
    \Block[l]{1-9}{\textbf{Qwen3.8-Max}} \\
    LLM + solver               & 162 & 6 & 55 & 76 & 20 & 3 & 0 & 2 \\
    SMART-LLM-style            & 132 & 9 & 30 & 46 & 31 & 14 & 0 & 2 \\
    RoCo-style                 & 187 & 4 & 46 & 87 & 30 & 18 & 0 & 2 \\
    HMAS-2-style               & 120 & 15 & 13 & 38 & 20 & 28 & 0 & 6 \\
    \midrule
    \textbf{\Scissors (ours)}  & 70 & 0 & 22 & 19 & 11 & 16 & 0 & 2 \\
    \midrule
    \Block[l]{1-9}{\textbf{Claude Sonnet 5.5}} \\
    LLM + solver               & 105 & 8 & 44 & 28 & 16 & 5 & 0 & 4 \\
    SMART-LLM-style            & 88 & 10 & 31 & 23 & 15 & 8 & 0 & 1 \\
    RoCo-style                 & 114 & 6 & 38 & 25 & 27 & 10 & 0 & 8 \\
    HMAS-2-style               & 102 & 18 & 13 & 27 & 21 & 17 & 0 & 6 \\
    \midrule
    \textbf{\Scissors (ours)}  & 42 & 0 & 14 & 4 & 12 & 9 & 0 & 3 \\
    \bottomrule
  \end{NiceTabular*}
\end{table*}

\begin{table}[t]
  \caption{\textbf{\Scissors minus the best other method, by part.} In points, paired by scene against the method with the highest accuracy in each part, with 95\% bootstrap intervals that resample the 16 task--level pairs (10\,000 resamples); a point is the difference of the two accuracies as printed in the main text and in \cref{tab:supp-robust}.}
  \label{tab:supp-cells}
  \centering
  \footnotesize
  \setlength{\tabcolsep}{2pt}
  \begin{NiceTabular*}{\linewidth}{@{\extracolsep{\fill}}lcc@{}}[cell-space-limits=1pt]
    \toprule
    Part & Qwen3.8-Max & Claude Sonnet 5.5 \\
    \midrule
    \ver{exact}     & $-$2.0 [$-$8.3, 4.2] & 2.1 [$-$2.1, 6.2] \\
    \ver{surplus}   & 9.4 [2.1, 16.7] & 14.6 [6.2, 24.0] \\
    \ver{shortfall} & 7.3 [1.0, 15.6] & 3.1 [$-$1.0, 7.3] \\
    \ver{swapped}   & 18.7 [8.3, 29.2] & 20.8 [13.5, 28.1] \\
    \ver{borrow}    & 8.4 [$-$10.4, 26.0] & 7.3 [$-$6.2, 19.8] \\
    \midrule
    \Con            & 11.7 [2.9, 18.3] & 9.2 [2.9, 15.0] \\
    \Rea            & 9.1 [1.7, 15.8] & 10.0 [5.8, 14.2] \\
    \midrule
    easy            & 14.6 [7.5, 20.8] & 10.4 [3.3, 16.7] \\
    hard            & 5.9 [$-$2.5, 13.8] & 7.0 [1.2, 12.5] \\
    \bottomrule
  \end{NiceTabular*}
\end{table}

\begin{table}[t]
  \caption{\textbf{Success with the same right teams.} Success (\%) of \Scissors (Ours) and of each method (Method) on the $n$ feasible episodes where both name a right team.}
  \label{tab:supp-same}
  \centering
  \footnotesize
  \setlength{\tabcolsep}{1.5pt}
  \begin{NiceTabular*}{\linewidth}{@{\extracolsep{\fill}}l*{6}{c}@{}}[cell-space-limits=1pt]
    \toprule
    \Block[l]{2-1}{Method} & \Block{1-3}{Qwen3.8-Max} & & & \Block{1-3}{Claude Sonnet 5.5} & & \\
    \cmidrule(lr){2-4} \cmidrule(lr){5-7}
    & $n$ & Ours & Method & $n$ & Ours & Method \\
    \midrule
    LLM + solver             & 193 & 55.4 & 20.2 & 208 & 57.2 & 25.5 \\
    SMART-LLM-style          & 196 & 51.5 & 24.0 & 223 & 57.0 & 27.4 \\
    RoCo-style               & 184 & 54.3 & 17.9 & 204 & 56.9 & 24.0 \\
    HMAS-2-style             & 199 & 54.8 & 22.1 & 220 & 58.2 & 26.8 \\
    \bottomrule
  \end{NiceTabular*}
\end{table}

\begin{table}[t]
  \caption{\textbf{Held-out seeds.} The main runs (\%) on the 400 scenes of seeds 1--5 and on the 80 of seed 0; \emph{Feas.}, \emph{Inf.}: the gold answer is feasible or infeasible.}
  \label{tab:supp-split}
  \centering
  \footnotesize
  \setlength{\tabcolsep}{2pt}
  \begin{NiceTabular*}{\linewidth}{@{\extracolsep{\fill}}l*{6}{c}@{}}[cell-space-limits=1pt]
    \CodeBefore
      \rowcolor{tabGroup}{3,9}
    \Body
    \toprule
    \Block[l]{2-1}{Method} & \Block{1-3}{Seeds 1--5 (400)} & & & \Block{1-3}{Seed 0 (80)} & & \\
    \cmidrule(lr){2-4} \cmidrule(lr){5-7}
    & Feas. & Inf. & All & Feas. & Inf. & All \\
    \midrule
    \Block[l]{1-7}{\textbf{Qwen3.8-Max}} \\
    LLM + solver             & 72.3 & 56.4 & 66.8 & 73.1 & 46.4 & 63.8 \\
    SMART-LLM-style          & 73.1 & 72.9 & 73.0 & 69.2 & 71.4 & 70.0 \\
    RoCo-style               & 67.3 & 46.4 & 60.0 & 71.2 & 57.1 & 66.2 \\
    HMAS-2-style             & 74.2 & 75.7 & 74.8 & 71.2 & \textbf{85.7} & 76.2 \\
    \midrule
    \textbf{\Scissors (ours)}   & \textbf{83.8} & \textbf{90.7} & \textbf{86.2} & \textbf{82.7} & 78.6 & \textbf{81.2} \\
    \midrule
    \Block[l]{1-7}{\textbf{Claude Sonnet 5.5}} \\
    LLM + solver             & 75.0 & 84.3 & 78.2 & 76.9 & 78.6 & 77.5 \\
    SMART-LLM-style          & 79.2 & 86.4 & 81.8 & 78.8 & 85.7 & 81.2 \\
    RoCo-style               & 73.5 & 86.4 & 78.0 & 61.5 & 78.6 & 67.5 \\
    HMAS-2-style             & 76.2 & 82.1 & 78.2 & 75.0 & 92.9 & 81.2 \\
    \midrule
    \textbf{\Scissors (ours)}   & \textbf{87.7} & \textbf{97.9} & \textbf{91.2} & \textbf{88.5} & \textbf{96.4} & \textbf{91.2} \\
    \bottomrule
  \end{NiceTabular*}
\end{table}

\input{figures/supp/supp-pertask}

\paragraph{Intervals.}
Paired by scene, \Scissors minus each method is 9.5 to 24.4 points over all episodes and 26.1 to 41.7 at the task level, and every interval excludes 0 (\cref{tab:supp-ci}): \Scissors is ahead of each method on both backbones.
The narrowest margin is over SMART-LLM-style with Claude Sonnet 5.5, 9.5 points [5.4, 13.8].
Resampling the 16 task--level pairs instead of the 96 task-scenes, every interval also excludes 0, the lowest bound being 5.0 points over all episodes and 11.5 at the task level, and resampling the 8 tasks gives the same.
Against each method, \Scissors is ahead on all 8 tasks on both backbones (\cref{fig:supp-pertask}): a one-sided sign test over the tasks gives $p = 1/256 \approx 0.004$.

\paragraph{By part.}
\Cref{tab:supp-cells} gives \Scissors' lead over the best other method in each configuration, suite and level, with intervals over the task--level pairs.

\paragraph{Same right teams.}
On the feasible episodes where \Scissors and a given method both name a right team, \Scissors succeeds in 51.5 to 55.4\% with Qwen3.8-Max and 56.9 to 58.2\% with Claude Sonnet 5.5, the other method in 17.9 to 24.0\% and 24.0 to 27.4\% (\cref{tab:supp-same}).

\paragraph{The executor's ceiling.}
Handed only the gold team, the shared executor succeeds in 58.7\% and 60.3\% of the 312 feasible scenes (Oracle-team); with a right team, \Scissors' episodes succeed in 55.9\% and 58.8\%, the other methods' in 18.9 to 25.2\% and 23.3 to 27.5\%.
Outside \ver{exact}, Oracle-script plays a scripted gold route for the gold team.
By configuration, \Scissors' MSS-SR is 42.7 to 50.0\% with Qwen3.8-Max and 47.9 to 55.2\% with Claude Sonnet 5.5, the other methods' 8.3 to 21.9\% and 12.5 to 25.0\%.

\paragraph{Held-out seeds.}
\Cref{tab:supp-split} gives each row on the 400 scenes of seeds 1--5, where \Scissors leads the best other method by 11.4 and 9.4 points, and on the 80 of seed 0.

\paragraph{Robustness.}
Across levels no method moves by more than 7.5 points (\Scissors with Qwen3.8-Max, 89.2\% at easy and 81.7\% at hard), and \Scissors is first at both levels on both backbones (\cref{tab:supp-robust}); across rooms none moves by more than 5.7 points (LLM + solver with Qwen3.8-Max, \Scissors with Claude Sonnet 5.5), and \Scissors is first in every room.
On \ver{swapped}, the other methods are right about as often where an SO-101 is brought in, which always makes the task infeasible, as where another arm is (together 51.0\% and 56.2\% with Qwen3.8-Max, 71.4\% and 72.4\% with Claude Sonnet 5.5): they do not read the SO-101 as infeasible.
On \ver{borrow}, taking the larger object would be right where the prop is larger, wrong where the distractor is; the gap is $-$8.8 and 1.7 points for \Scissors and $-$27.6 to 5.0 for the others, so no method chooses by size.
Some answers contradict another on the same task-scene, whatever the gold: \ver{surplus} adds an arm to \ver{exact}, so it cannot be infeasible or need more arms where \ver{exact} is feasible; \ver{shortfall} takes one away, so it cannot be feasible where \ver{exact} is not, or with fewer arms; on Rea-02 and Rea-04, \ver{swapped} puts a Franka in a PiPER's place, so it cannot be infeasible or need more arms where \ver{exact} is feasible. Of these 216 pairs per method, \Scissors contradicts itself in 12 with Qwen3.8-Max and 3 with Claude Sonnet 5.5, the other methods in 29 to 60 and 16 to 25, and \Scissors never answers \ver{surplus} infeasible after a feasible \ver{exact}.

\paragraph{Per-task analysis.}
\Cref{fig:supp-pertask} shows where each method errs: a \ver{borrow} cell is the hardest, alone or tied, for every method on both backbones but RoCo-style with Qwen3.8-Max, whose hardest is Con-04 \ver{swapped} (8.3\%).
\Scissors is first on every task on both backbones, by 3.3 to 16.7 points with Qwen3.8-Max and 1.7 to 13.3 with Claude Sonnet 5.5; its lead is largest on Con-02 and Con-03 with Qwen3.8-Max (16.7 and 13.3) and on Rea-03 and Con-04 with Claude Sonnet 5.5 (13.3 and 11.7).

\paragraph{Errors by kind.}
\Scissors never names an arm too many, while the other methods do so 4 to 15 times with Qwen3.8-Max and 6 to 18 times with Claude Sonnet 5.5 (\cref{tab:supp-errors-main}); it gives a team to an infeasible task 19 and 4 times, against 38 to 87 and 23 to 28.
Its other errors are a missing arm (22 and 14), the distractor used (16 and 9), a feasible task refused (11 and 12) and another team of the fewest arms (2 and 3).

\paragraph{\Scissors' errors by configuration.}
\ver{borrow} holds most of \Scissors' wrong answers, 39 of 70 with Qwen3.8-Max and 32 of 42 with Claude Sonnet 5.5.
There it takes the distractor (16 and 9), names one arm with the prop where two are needed (15 and 11), or declares the task infeasible (8 and 12).
With Qwen3.8-Max, 16 more are on \ver{swapped}, 13 of them a team given where the swap leaves a job without a doer.

\subsection{Prompts}
\label{sec:supp-prompts}

Each model session opens with a prompt as its system message: a line naming the session's role, the shared part where the prompt includes \texttt{\{COMMON\}}, and the method's own instructions.
We give key passages of the shared part and of each method's own instructions, verbatim; \texttt{[...]} marks omitted text, a bracketed label names the session and is not sent to the model, and the full prompts will be released with the code.
Braced fields are filled before the call: \texttt{\{ARM\}} and \texttt{\{ARMS\}} with an arm's name and the team's, \texttt{\{SIDE\}} with one of a reviewer's two sides.
Images reach a session only through \texttt{look}, and the tools are those of \cref{tab:supp-tools}.
The prompts call passive objects \emph{things}.

\lstdefinestyle{mrprompt}{
  basicstyle=\ttfamily\scriptsize,
  columns=fullflexible,
  keepspaces=true,
  breaklines=true,
  breakatwhitespace=true,
  breakindent=8.4pt,
  showstringspaces=false,
  literate={`}{{\textasciigrave}}1 {'}{{\textquotesingle}}1 {"}{{\char34}}1,%
  aboveskip=0pt,
  belowskip=0pt,
  tabsize=2
}

\newtcolorbox{mrpromptbox}[1]{
  breakable,
  title={#1},
  fonttitle=\footnotesize\bfseries,
  colback=white,
  colframe=gray,
  coltext=black,
  boxrule=0.6pt,
  left=1mm,
  right=1mm,
  top=1mm,
  bottom=1mm
}

\paragraph{Shared part.}
It states the question and the form of the answer, the same for every method.

\begin{mrpromptbox}{Shared part (\texttt{\{COMMON\}})}
\begin{lstlisting}[style=mrprompt]
## The question

A team of robot arms is mounted round one table, with things on it and a goal. The question is how big a team the goal needs:

- which arms are needed -- the fewest that can do the whole task -- and which can stay idle;
- whether a thing on the table can stand in for an arm [...];
- or that the task cannot be done with what is here, and what is missing.

## The answer: `submit_team_plan`

- `feasible`; `arms_used` (the fewest arms that do it; empty if it cannot be done); `idle` (every other arm, each with why);
- `things_used` (each with the job it does and, if it takes over an arm's job, which); [...]
- `missing` (if it cannot be done: which job has no doer, and what kind of arm or thing would do it);
- `steps` (who does what, and what happens at the same time); `key_constraints` (what forces the number of arms, with numbers).
[...]
\end{lstlisting}
\end{mrpromptbox}

\paragraph{Baselines.}
SMART-LLM-style runs its three stages as three sessions, each receiving the output of the stages before it.
In RoCo-style each arm speaks in a session of its own, in the order \W, \Ss, \E, \N of the arms present, for at most three rounds, and receives the dialogue so far and the latest proposal.
In HMAS-2-style the central planner receives the plan board, its earlier proposals with the arms' answers, and each arm checks the new proposal in a session of its own, for at most three rounds.
LLM + solver is one session.

\begin{mrpromptbox}{Baselines: own instructions}
\begin{lstlisting}[style=mrprompt]
[SMART-LLM-style / decomposition]
Look at the scene (`briefing`, `look`, `locate`, `measure`) and break the goal into subtasks. [...]
Do not assign arms yet.

[SMART-LLM-style / coalition formation]
For each subtask, find who can do it:
- **arms**: query each arm about the subtask's skills -- `can_reach` at its place, pointing the way it needs (read the margin), `read_arm` for the fingers' opening, `heft` where a mass decides it;
- **things** on the table that could do it instead of an arm [...]: `measure` them and say why each would or would not do.
A subtask that no arm and no thing can do gets an empty coalition.

[SMART-LLM-style / task allocation]
Allocate each subtask to a member of its coalition, using the minimum number of robots, so that subtasks that must happen at the same time get different doers and every subtask's order is kept. [...] If some subtask's coalition is empty, the task cannot be done: say which subtask, and what kind of arm or thing would do it.

[RoCo-style / each arm]
You are the {ARM} arm, one of the robot arms ({ARMS}) mounted round one table, each run by its own copy of you [...]. You talk with the other arms to agree on a team plan; nothing moves until all of you agree on one.
[...]
You can query only your own joints (`read_arm`) and your own kinematics (`can_reach`); ask the others about theirs in your message.
[...]
- `data.taking_part`: whether you take part (you may choose not to);
[...]
The dialogue ends when every arm agrees with one proposal.

[HMAS-2-style / central planner]
You propose the whole plan; each arm checks its own part with its own kinematics and accepts or vetoes it. You do not query the arms' kinematics yourself [...] If the plan board shows vetoes, revise the plan to answer them.

[HMAS-2-style / each arm]
Check your own part with your own tools -- `can_reach` at the places it asks you to work (read the margin), `read_arm`, [...] -- or, if it leaves you idle, whether that is right. Then `respond`: accept or veto (`data.accept`), the reason, and your evidence (the tool results).

[LLM + solver]
You translate the task into a job graph; a solver counts the arms [...].
[...]
2. **Write the jobs.** Where a hand (or a thing) must be and what it must do, with the force and for how long. Join with "with" the jobs that must happen at the same time. [...]
3. **Find the doers, with evidence.** [...] A thing is a doer of a job only with evidence that it does it [...] and some arm must put it in place first: add that job.
4. **Solve.** `solve_team` with the jobs and transfers. If it is blocked, read why; fix the graph only with evidence and solve again. If nothing helps, the task cannot be done: say what is missing.
\end{lstlisting}
\end{mrpromptbox}

\paragraph{\Scissors.}
The prover receives the briefing and the scene graph without the arms and, of the hands, only the widest opening; it has no reach tool, and its camera pictures show the whole table, arms included. The stand-in pass receives the job graph and the jobs without a doer; an object it adds comes with the job of placing it, and the jobs already written stay as they are.
The two reviewers share one prompt, with \texttt{\{SIDE\}} filled by one of two sides, and receive the goal, the job graph, the certificate and the flip-tested elements; the arbiter receives the goal, the job graph and the edits in dispute.

\begin{mrpromptbox}{\Scissors: own instructions}
\begin{lstlisting}[style=mrprompt]
[prover]
You model a task on a table as a **job graph**: what the goal asks to be done, where, and what must happen at the same time. Robot arms will do it, but you are not told which arms there are, where they stand or what they reach -- that is measured afterwards, by code, for every arm at every place you name. [...]
- **jobs** -- what must be done, and where a hand (or a thing standing in for one) must be to do it. [...]
- **group** -- jobs with the same group name are done at the same time, each by its own hand. [...]
- **things** -- a body on the table that is not what the goal names, which the plan puts in place so that the task works [...] list every candidate under `things` with the sizes that decide, and code measures each and keeps the one that fits. [...]
Write every job the goal needs; one the goal does not need is a mistake too.

[stand-in pass]
Your one question now: **is there a body on the table that would do any of these jobs once put in place?** [...] The jobs themselves stay as you wrote them: you cannot change, drop or add a job here. [...]

[reviewer]
Every cell (who reaches where, who meets whom) is a tool result. [...] What may be wrong is what the model wrote: a job missing or not needed, a place written where it is not, what is or is not done at once, who may do a job, a thing's spec. [...]
{SIDE}
[...]
Each edit needs `evidence`: the tool result it rests on, **from your own calls in this session**; an edit whose evidence names no tool you called is refused. An edit that makes the task easier also needs `quote`: the words of the goal it rests on, copied exactly.
[...]
Code applies each edit, measures what it names and draws the answer again. An edit that changes nothing is dropped. One that makes the task harder for the arms is kept on its evidence. One that makes it easier is not kept on its own: it goes, with its quote and evidence, to an arbiter, who keeps it only if the goal and the scene bear it out.

[{SIDE}: too easy]
Look for what the graph leaves out or writes easier than the task is:
- a job the goal needs that is missing [...];
- jobs that must be done at once and are not in one group. [...]

[{SIDE}: too hard]
Look for what the graph asks that the goal does not:
- a job the goal does not need;
- jobs in one group that need not be done at once (one could be done and let go before the other);
[...]
- a job no hand can do that a thing on the table would do once put in place [...]

[arbiter]
For each unit of edits, decide one thing: does the goal, read as written, and the scene, as the tools show it, bear the edit out? Keep it only if they do. If you cannot tell, refuse it: the model as written stands.
\end{lstlisting}
\end{mrpromptbox}

\paragraph{Shared executor and form check.}
The executor writes the program for a method's team plan with the program API of \cref{tab:supp-tools}.
A plan that RoCo-style or HMAS-2-style submits and that fails the form check goes once to a session that fixes only its form.

\begin{mrpromptbox}{Shared executor and form check}
\begin{lstlisting}[style=mrprompt]
[executor]
You write the program that a team of robot arms mounted round one table runs to do a task. Which arms take part, which stay idle and which things on the table are used is already decided: the team plan, in the first message. [...]
[...]
Keep to the team plan: do not move an arm it leaves idle.

[form check]
You put an agreed team plan into the submission form. [...] Do not change the decision.
\end{lstlisting}
\end{mrpromptbox}

\end{document}